\documentclass[journal]{IEEEtran}
\usepackage{fix-cm}
\usepackage{xcolor,soul,framed} 
\colorlet{shadecolor}{yellow}
\usepackage[pdftex]{graphicx}
\graphicspath{{../pdf/}{../jpeg/}}
\DeclareGraphicsExtensions{.pdf,.jpeg,.png}

\usepackage[cmex10]{amsmath}
\usepackage{amssymb}
\usepackage{array}
\usepackage{algorithm}

\usepackage{eqparbox}
\usepackage{url}
\usepackage{verbatim}
\usepackage{amsmath}
\usepackage{tabularx}
\usepackage{amsthm}
\usepackage[english]{babel}

\usepackage{booktabs}
\usepackage{multirow}
\usepackage{algpseudocode}

\usepackage{tikz}
\usetikzlibrary{arrows.meta,positioning}

\usepackage{xcolor} 
\usepackage{pifont} 

\newcommand{\cmark}{\textcolor{green!80!black}{\ding{51}}}
\newcommand{\xmark}{\textcolor{red}{\ding{55}}}

\usepackage{array,tabularx,longtable,booktabs}
\newcolumntype{L}[1]{>{\raggedright\arraybackslash}p{#1}}
\newcolumntype{Y}{>{\raggedright\arraybackslash}X}

\usepackage{xr}

\def\showrevision{}
\usepackage{xcolor}
\usepackage[normalem]{ulem}       
\usepackage{soul}        
\usepackage{marginnote}  
\usepackage{xspace}

\definecolor{revnewcolor}{RGB}{230,115,0}   
\definecolor{revoldcolor}{RGB}{107,142,35}  
\definecolor{revdelcolor}{RGB}{80,80,80}    
\definecolor{revmodcolor}{RGB}{0,0,200}     
\definecolor{revrefcolor}{RGB}{180,0,180}   

\ifdefined\showrevision

  \newcommand{\revold}[1]{{\color{revoldcolor}#1}}

  \newcommand{\revdel}[1]{{\color{revdelcolor}\sout{#1}}}

  \newcommand{\revref}[2]{%
    \marginnote{\color{revrefcolor}\footnotesize$[\text{R}_{#1}\text{C}_{#2}]$}%
  }

\ifdefined\showrevision
  \newcommand{\robustrevref}[2]{{\color{revrefcolor}\footnotesize$[\text{R}_{#1}\text{C}_{#2}]$}}
\else
  \newcommand{\robustrevref}[2]{}
\fi

\else

  \newcommand{\revold}[1]{}            
  \newcommand{\revdel}[1]{}            
  \newcommand{\revref}[2]{}            
  \newcommand{\robustrevref}[2]{}      

\fi

\newif\ifblind
\blindfalse 

\begin{document}
\bstctlcite{IEEEexample:BSTcontrol}
    \title{Learning From a Steady Hand: A Weakly Supervised Agent for Robot Assistance under Microscopy}

\ifblind

    \author{Anonymous Authors} 
\else
  \author{Huanyu Tian$^{1}$,
      Martin~Huber$^{1}$,
      Lingyun~Zeng$^{1}$,
      Zhe~Han$^{1}$,
      Wayne~Bennett$^{2}$,
      Giuseppe~Silvestri$^{2}$,  
      Gerardo~Mendizabal-Ruiz$^{2,4}$, 
      Tom~Vercauteren$^{1}$, 
      Alejandro~Chavez-Badiola$^{3,2}$,
      and~Christos~Bergeles$^{1,2}$

\thanks{*This work was supported by Innovate UK under grant agreement 10111748. Accepted for publication in IEEE Transactions on Robotics (T-RO), 2026.}
\thanks{$^{1}$H.~Tian, M.~Huber, Z.~Han, L.~Zeng, T.~Vercauteren, and C.~Bergeles are with the School of Biomedical Engineering \& Imaging Sciences, King’s College London, UK. Corresponding author: {\tt\small huanyu.tian@kcl.ac.uk}}    
\thanks{$^{2}$W.~Bennett, G.~Silvestri, G.~Mendizabal-Ruiz, C.~Bergeles, and A.~Chavez-Badiola are with Conceivable Life Sciences, New York City, US and London, UK.}
\thanks{$^{3}$A.~Chavez-Badiola is with Hope IVF Mexico, Mexico.}
\thanks{$^{4}$ G.~Mendizabal-Ruiz is with Departamento de Bioingeniería Traslacional, Universidad de Guadalajara, Guadalajara, Jalisco, Mexico}
}
\fi

\ifblind
\markboth{
}{ \MakeLowercase{\textit{et al.}}: Learning From a Steady Hand: A Weakly Supervised Agent for Robot Assistance under Microscopy}
\else
\markboth{
}{Tian \MakeLowercase{\textit{et al.}}: Learning From a Steady Hand: A Weakly Supervised Agent for Robot Assistance under Microscopy}
\fi

\maketitle


\begin{abstract}
This paper rethinks steady-hand robotic manipulation by using a weakly supervised framework that fuses calibration-aware perception with admittance control. 
Unlike conventional automation that relies on labor-intensive 2D labeling, our framework leverages reusable warm-up trajectories to extract implicit spatial information, thereby achieving calibration-aware, depth-resolved perception 
without the need for external fiducials or frame-by-frame
manual depth annotation.
By explicitly characterizing residuals from observation and calibration models, the system establishes a task-space error budget from recorded warm-ups.
Under the evaluated laboratory conditions, the recorded trajectories yielded an empirical lateral error level of approximately $49\,\mu$m and a conservative axial error estimate of approximately $291\,\mu$m during large in-plane moves.
In a within-subject user study \((N{=}8)\), the learned agent reduces overall NASA-TLX workload by \(77.1\%\) relative to the simple steady-hand assistance baseline. 
Together, these results demonstrate successful closed-loop micromanipulation and reduced subjective workload under the evaluated laboratory conditions, without external fiducials or frame-by-frame manual depth annotation, supporting the proposed approach as a framework for microscope-guided robotic assistance.

\end{abstract}


%
\IEEEpeerreviewmaketitle


\section{Introduction}
\label{sec:intro}

\IEEEPARstart{R}{obotic} micromanipulation \cite{zhangmicromanipulation}, in this work, is related to offering robotic assistance to a user that operates under microscopic imaging to repeatedly bring cells or other micro-objects into view and perform sampling (pickup/placement). 
We are specifically interested in micromanipulation processes related to egg/embryo vitrification~\cite{embryovirtrification, boylancell, changvitrification}.
Oocyte and embryo vitrification outcomes are sensitive to operator dexterity and consistency, motivating the development of robotic assistance that can reduce inter-operator variability. Physiological hand tremor alone introduces positional instability of $\delta_{\mathrm{tremor}}\!\approx\!202\,\mu$m during station-keeping tasks~\cite{riviere1997characteristics}, larger than the target cell diameter ($100$--$150\,\mu$m). Beyond tremor, the procedure demands precise cryoprotectant exposure timing and consistent tool approach angles, both of which are highly sensitive to inter-operator variability.
Addressing this clinical need requires a system that operates across two distinct scales:
a macro layer operating within a decimeter-scale workspace for tool exchange and cross-dish transport, and a micro layer executing precision tasks within the centimeter-scale visual field. Given biological targets (e.g., oocytes or embryos) ranging from $100$ to $150\,\mu\mathrm{m}$, and accounting for the hydraulic aspiration, the system requires a positioning precision strictly within sub-millimeter—with higher accuracy being increasingly advantageous for safety~\cite{tan2007monodisperse},~\cite{guevorkian2017micropipette}.

\begin{figure}[t]
    \centering
    \includegraphics[width=1.0\columnwidth]{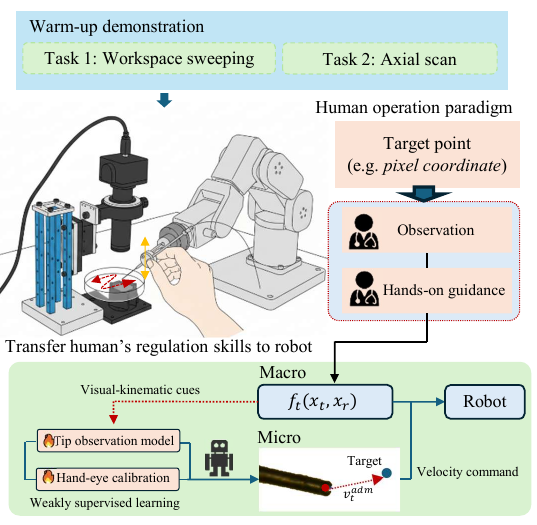}
    \caption{
    The data collection relies on ``warm-up'' episodes: short trajectories generated by the operator dragging the tool under steady-hand guidance. These maneuvers are standardized into two tasks: lateral sweeps for FOV coverage (the red line), and axial scans for controlled depth-axis motion (the orange line).
    The agent leverages warm-up trajectories to train the 3D tip observer and estimate hand-eye fusion parameters. Leveraging the dataset-learned calibration and tip recognition, the agent servos laterally to the operator-designated target (e.g., for aspiration) and regulates axial depth for station-keeping.
    }
    \label{fig:overall}
\end{figure}

In clinical practice, the end-effector assembly comprises a rigid metal holder and a replaceable stripper tip (a flexible medical consumable) used to aspirate and transport cells~\cite{chavez2025automated}.  Frequent tool replacement introduces inevitable assembly variances,  rendering the tip-to-base kinematics unreliable.  This  kinematic incompleteness, combined with strict sterilization and disposability requirements that preclude permanent marker attachment, motivates our core design principle: rather than relying on task-level demonstrations, the  system infers tip kinematics from non-task-specific warm-up maneuvers comprising lateral workspace sweeps and axial depth scans,  making the framework deployable without prior knowledge of tool geometry.

The resulting system presents the following interaction paradigm to the operator: the operator implicitly adjusts the target depth by focusing the microscope and designates the lateral position via a screen click; the robotic agent then autonomously servos the micropipette tip to this 3D setpoint for precise aspiration or dispense.
Given that the dynamics of the specimen are much slower than the robot's control loop frequency (i.e., control bandwidth), the target is modeled as a static setpoint. This simplifies the control task to point-to-point regulation rather than trajectory tracking.

\subsection{Hierarchical Technical Challenges}

Recent learning-based approaches also face a common challenge: the lack of a scalable supervision source compatible with transparent tools, monocular optics, real-time constraints, and marker-free deployment.
The data-efficient methods such as Diff-DAgger~\cite{lee2025diff} and diffusion-policy-based approach (e.g. SKIL)~\cite{Wang2025SKILSK} reduce demonstration requirements, but share a critical assumption: that  task-level demonstrations can be collected. In our setting, this assumption is circular: without robotic assistance, operators 
struggle to perform the precise, repeatable demonstrations required  to train the system that would provide that assistance. 
Consequently, task-level demonstrations are not merely scarce in this domain, they are structurally infeasible without the very system we seek to build.
These constraints give rise to a hierarchy of interconnected technical challenges:
\begin{table}[t]
\footnotesize  
\centering
\caption{Comparison of Features across Different Micromanipulation Paradigms.}
\label{tab:micromanipulation_comparison}
\begin{tabularx}{\columnwidth}{l Y Y Y}
\toprule
\textbf{Features} & \textbf{Visual \newline Servoing} & \textbf{Stereo-\newline tactic} & \textbf{Steady-Hand} \\
\midrule
\textit{Perception} & & & \\
\quad Markerless & \cmark & \xmark & \cmark \\
\quad Occlusion robust & \xmark & \cmark & \cmark \\
\quad Calibration-free & \cmark & \xmark & Few warm-ups \\
\midrule
\textit{Control} & & & \\
\quad Macro-to-micro & \xmark & \cmark & \cmark \\
\quad Re-entry recovery & \xmark & \cmark & \cmark \\
\quad 3D servoing & \xmark & \cmark & \cmark \\
\midrule
\textit{Deployment} & & & \\
\quad Few-prompt annotation & \xmark & \cmark & \cmark \\
\quad Rapid adaptation & \xmark & \xmark & \cmark \\
\bottomrule
\end{tabularx}
\end{table}

\noindent \textbf{Microscopy imaging}: 
Typically fabricated from glass or polymer, the tools appear partially transparent, specular, and refractive.
These optical properties violate the assumptions of conventional edge or texture detection, and represent a unique challenge: the optical path traverses an air-liquid interface, introducing significant spherical aberration that makes defocus patterns far more complex than in air.
To address this, classic pipelines generally localize the tip by relying on hand-crafted motion heuristics 
(e.g., background subtraction or optical flow) rather than static shape priors.

Sharing similar instrument tracking challenges, the field of  endoscopic/laparoscopic surgery has seen progress with datasets like ROBUST-MIPS~\cite{han2025robust} that provide instance masks and skeletal poses. Building on these, recent methods estimate 2D keypoints using vision-language models or low-rank adaptation~\cite{duangprom2025estimating}, often strengthened by multi-frame context.
Such pipelines are resilient to occlusions and lighting variability, and leverage benchmarks to standardize 3D pose evaluation~\cite{XU2025103674}. Directly translating them to microscopy, however, exposes new gaps: 
\begin{itemize}
    \item \textbf{Domain shift:} endoscopic tools possess rich texture, whereas transparent micropipettes lack the features required for standard descriptors.
    \item \textbf{Latency:} state-of-the-art foundation backbones (e.g., SAM/ViT~\cite{ravi2024sam2}) require hundreds of milliseconds per frame, stressing $\ge$30\,Hz control loops.
    \item \textbf{Control:} most pipelines output 2D masks/keypoints, and not 3D tip states as required for micro-scale manipulation.
\end{itemize}

Crucially, since the keypoint detector functions as the feedback sensor within the visual servoing loop, its inference latency manifests as a pure time delay that directly erodes the system's phase margin. Therefore, provided that detection accuracy is maintained, minimizing inference time is paramount to maximize the bandwidth and stability of the closed-loop control.

\noindent \textbf{Depth estimation}: 
Surgical videos often leverage stereo geometry~\cite{allan2021stereo}. The photo-consistency assumption required for stereo correspondence is often violated in microscopy when observing transparent or specular surfaces, leading to reconstruction artifacts. Thus, the system has to rely on monocular defocus, whereas microscopes only have local and weak depth cues~\cite{Bud_Transferring_MICCAI2024}.
To estimate depth, most approaches rely on defocus-based focal stacking or iterative focus scans~\cite{zhang2023sift, liu2013tip}.
While robust, these procedures are slow and inherently local, constrained by directional ambiguity (due to symmetric blur) and the fact that defocus gradients vanish far from the focal plane~\cite{gur2019single}.
Monocular depth estimation under microscopy has also been explored, for example z-axis focus sweeps to localize the region of interest (ROI)~\cite{wang2023Grasp}, depth detection via contact observation~\cite{yang2024depth}, and pure defocus estimation networks~\cite{luan2025autofocusing}. However, these methods typically rely on synthetic Gaussian blur to model the defocus-depth relationship. 
This assumption proves inadequate in micromanipulation: as the transparent tip traverses the air-liquid interface under concentrated illumination, it generates complex, non-uniform specular reflections and refractive artifacts that fundamentally deviate from simple Gaussian models, rendering such synthetic mappings ineffective. 
Overcoming this limitation via supervised learning is difficult because annotated microscopy datasets are scarce: ambiguous boundaries, glare, and tool translucency inflate labeling cost.

\noindent \textbf{Visual-motor policy under the microscope}:
To achieve micron-level precision, micromanipulation systems commonly follow three paradigms in Table.~\ref{tab:micromanipulation_comparison} with distinct trade-offs among perception, control, and deployment.
By maintaining continuous feature tracking strictly within the microscope's field of view (FoV), calibration-free visual servo methods can achieve pixel-level lateral precision\footnote{Lateral precision refers to positioning error on the imaging plane.} without explicit markers, offering high-precision assistance during the micro-manipulation phase~\cite{Yang2020hybrid}.

Stereotactic approaches operate within a globally calibrated framework, coupling accurate robot kinematics with 3D sensing of robot links via external imaging system~\cite{lu2023markerless,huber2025hydra}. These approaches provide macro workspace coverage and globally observable depth. Typical lateral precision  is sub-millimetric ($0.5$-$1\,$mm)~\cite{Spyrantis2022ROSA}. These systems generally require explicit kinematics/hand-eye calibration and are sensitive to tool changes, for which recalibration is commonly performed \cite{liu2024gbec}. 

Steady hand fuses the capability of practitioners with robotic control, offering coverage and $0.03$-$1\,$mm lateral precision. 
Depth control is locally reliable via defocus cues localized around the tool tip.
A short warm-up trajectory (routine calibration maneuvers) suffices to enter stable closed-loop control, and the scheme is marker-free and tolerant to tool exchange. However, this paradigm imposes a sustained cognitive load: the operator must continuously mentally map the visual error (between the tool tip and the target) into physical guiding forces, a process that becomes increasingly fatiguing during repetitive high-precision tasks~\cite{tian2024semiautonomouslaparoscopicrobotdocking}~\cite{steadyhandaccuracy}.

\noindent \textbf{Macro-micro transition and confidence-gated safeguards}: 
Micromanipulation must smoothly shift from macro positioning to micro precision under marker-free, time-limited conditions. 
Rather than relying on external-sensor-based drift-prone localization of microscope/tool tip, steady-hand co-manipulation uses interaction-force-driven admittance control to let the operator guide the tool into view with tremor suppression, providing tremor-suppressed macro-level guidance and workspace coverage~\cite{tian2025uncertainty}. 

Given the inevitability of uncalibrated starts and visual tracking failures (e.g., due to occlusion or low-contrast microscopic imagery), a pivotal challenge lies in managing the reliability gap between automated servoing and human perception. The system cannot rely strictly on visual servo; instead, it faces the requirement of hierarchical authority allocation. Specifically, it must limit autonomous interventions to high-confidence scenarios while guaranteeing that the operator retains continuous manual override capability (i.e. the operation mode in macro layer) as a seamless fallback.

\subsection{Learning from a steady hand}
We resolve this infeasibility by reframing the data collection problem: instead of requiring task-level demonstrations, we exploit the steady-hand co-manipulation paradigm~\cite{SteadyHand1, steadyhandheld} as a source of non-task-specific supervision, where the robot and user simultaneously hold the instrument while an admittance controller maps interaction forces into tremor-suppressed end-effector motion~\cite{sharedcontrolsafety}.

Let $\mathbf{f}_t$ be the measured tool-handle force, $\mathbf{x}_t$ the target pose, $\mathbf{x}_r$ the actual pose, and $\mathbf{v}^{\mathrm{adm}}_t$ the velocity command. A standard admittance law:
\begin{equation}
\mathbf{M}_d\,\dot{\mathbf{v}}^{\mathrm{adm}}_t + 
\mathbf{B}_d\,\mathbf{v}^{\mathrm{adm}}_t = 
\mathbf{f}_t(\mathbf{x}_t,\mathbf{x}_r),
\label{eq:admittance}
\end{equation}
generates a compliant motion (with virtual inertia $\mathbf{M}_d$ and damping $\mathbf{B}_d$) toward the user's intent $\mathbf{x}_t$. During warm-up, the operator physically guides the tool under steady-hand guidance, generating kinematic-visual streams that serve as supervision for the perception and calibration modules.

Once trained, the system transitions to autonomous micro-level control. However, constructing a 3D target $\mathbf{x}_t$ is infeasible under monocular-microscope imaging guidance due to partial observability (missing depth) and kinematic uncertainties. To resolve this, we bypass explicit position planning and directly synthesize the guidance velocity $\mathbf{v}^{\mathrm{adm}}_t$ via calibration-aware perception:the operator designates a 2D target to generate lateral velocity, while a learning module infers axial velocity from defocus cues~\cite{luan2025autofocusing}; this composite velocity is then transformed via calibration matrices to form the global control input.

We do not consider ``learning from a steady hand'' as end-to-end policy learning. Instead, we treat the short co-manipulation warm-ups (cf. Fig.~\ref{fig:overall}) as a weak teacher (weakly-supervised learning~\cite{weakteacher1, weakteacher2}), where image-motion cues provide supervision for both spatial awareness and visual representation.

This warm-up-based supervision strategy has two practical consequences. First, warm-up trajectories can be collected in minutes under steady-hand guidance without task expertise, supporting rapid setup-specific initialisation without requiring task-level demonstrations.  Second, the hand-eye calibration module serves a dual role: beyond estimating the camera-to-robot rotation, the resulting registration error provides an automatic consistency check on the pseudo-label quality.  If the 2D tip trajectory and the 3D robot motion are geometrically self-consistent, the pseudo-labels are reliable; clips that fail this check can be flagged for targeted human review, replacing exhaustive manual review with a principled, error-driven audit.

\subsection{Contributions}

Consequently, we propose a weakly supervised framework that learns lateral/depth models and derives calibration parameters directly 
from visuo-kinematic cues during warm-ups. This manuscript  substantially extends~\cite{tian2025uncertainty} toward a complete  weakly supervised learning framework. While the preliminary work  was limited to planar (2D) control relying on fiducial markers, 
this extension incorporates axial depth regulation and integrates  uncertainty estimation into the control loop.

The principal contribution is a unified warm-up-based supervision workflow for medical micromanipulation. The framework uses non-task-specific visual--kinematic trajectories as a shared source for perception learning, relative depth estimation, markerless registration, and control initialization. This workflow combines established components into an experimentally evaluated robotic system without requiring task-level demonstrations or external markers. It is realised through three connected components:

\begin{itemize}
    \item \textbf{Warm-up-supervised 3D perception:} A two-stage tip observer that learns lateral position and relative depth from the visual--kinematic cues collected during warm-up, providing 3D feedback for closed-loop regulation.   Estimates are fused via a filter to ensure robust tracking under occlusions.

    \item \textbf{Warm-up-driven markerless calibration:} A visual--kinematic calibration component that reuses the warm-up trajectories to align perception with robot control and assess the consistency of the weak supervision, without external markers or prior tool geometry.

    \item\textbf{Warm-up-enabled adaptive shared control:} A hierarchical controller that uses the learned 3D tip feedback and setup-specific registration for confidence-gated micro-scale regulation, while retaining steady-hand guidance and manual authority at the macro scale.
\end{itemize}

We conducted a preliminary user study ($N=8$) comparing manual, steady-hand, and the proposed shared-control modes.
Results show that the Agent condition reduced self-reported overall NASA-TLX workload by approximately $77.1\%$ relative to steady-hand control. This provides preliminary evidence that the framework reduces the subjective workload gap between novice and experienced operators.

\section{Learning from A Steady-hand: Few-Prompt annotation}
\label{sec:pseudoLable}

This section reframes tip localization as learning from steady-hand motion feedback.
Visual features and hand-eye spatial relationships are extracted from the two kinds of dedicated warm-up demonstrations. We then use defocus to generate approximate depth labels, forming a defocus-derived depth dataset. 

\begin{figure}[t]
    \centering
    \includegraphics[width=1.0\columnwidth]{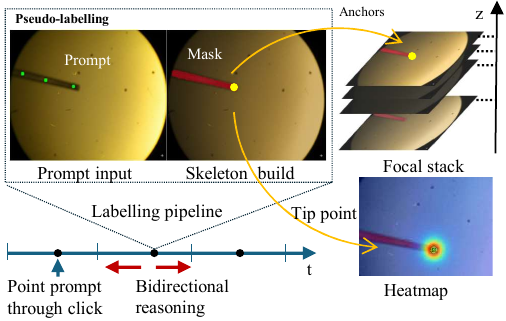}
    \caption{
     To circumvent dense manual labeling, we utilize ``steady-hand'' warm-up clips. By manually prompting SAM 2 on only a few anchor frames (sparse annotation), the system propagates tip masks bidirectionally to generate dense, real-time 3D tip estimation targets.
    }
    \label{fig:vision}
\end{figure}

\subsection{Data Accumulation from Warm-ups}
While the micropipette tool is frequently replaced in experiments, a ``setup change'' occurs only when the relative orientation between the robot and the microscope is altered. Therefore, a warm-up sequence is mandatory only for these new setups to establish the hand-eye calibration.

We leverage this calibration routine to automatically generate a labeled dataset of synchronized images and poses initiated by sparse human prompts. These warm-up samples including robot states and images are accumulated and highly reusable: the consistent visual appearance and defocus-kinematics patterns are utilized to train 3D tip depth estimation, while the established kinematics-vision correspondence is leveraged to calculate hand-eye calibration to transform velocity command from camera frame to the robot frame. These robust vision models subsequently enable the real-time feedback controller detailed in Section~\ref{sec:micro_controller}.

\subsection{Generation of Micropipette Tip Pseudo-Labels}

Manual annotation of tool tips in microscope videos is time consuming and inconsistent due to the changing visual appearance of the tool tip and the effect of defocus. 
Our proposed strategy leverages zero-shot segmentation models (e.g. SAM2~\cite{ravi2024sam2}) but addresses their memory-intensive shortcoming by restructuring long-form microscope sequences into uniformly sampled, fixed-length clips (Fig.~\ref{fig:vision}). Each clip receives prompt inputs propagated from reference frames where labeling is done manually. Within a clip, segmentation masks are propagated bidirectionally.

Given that full segmentation is both costly and unnecessary at inference, we extract the tool tip as a compact geometric feature via skeletonization. From the skeleton, we extract the set of endpoint candidates \(E_{t}\), defined as pixels having a single neighbor.

The tip identification is formulated as a \textbf{selection task} maximizing a multi-criteria score \(s_t(\mathbf{x})\) over \(\mathbf{x} \in E_{t}\).
The tip selection employs a score that balances spatial and temporal cues. While the entry point is inherently located at the image boundary, the tool tip may also approach the boundary during peripheral manipulation tasks. In such scenarios, spatial priors alone (e.g., proximity to the image center) are insufficient to distinguish the true tip from the entry point. To resolve this ambiguity, we incorporate temporal motion consistency.
For each frame \(t\), let \(E_{t}\) be the set of endpoint candidates (each \(\mathbf{x}_{\mathrm{ca}}=(x_1,x_2)\in E_{t}\)), and let \(\tilde{\mathbf{p}}_{\mathrm{tip},t-1}\) and \(\tilde{\mathbf{p}}_{\mathrm{tip},t+1}\) denote the provisional tip estimates obtained from a forward and a backward pass, respectively (bidirectional inference). 
We define spatial proximity as being away from the image boundary: since the other physical endpoint is the entry point, which typically lies on or very near the boundary, the true tip should be the endpoint farther from the boundary. 
This acausal design makes the pointwise score \(s_t(\mathbf{x}_{\mathrm{ca}})\) depend on both the previous and the next frames’ provisional tips. The border distance is:
\begin{equation}
    d_{\mathrm{border}}(\mathbf{x}_{\mathrm{ca}})=\min\{x_1,\;W-x_1,\;x_2,\;H-x_2\}.
\end{equation}
where \((W,H)\) denote the image width and height, respectively. A normalization can be applied to the border distance:
\[
\tilde d_{\mathrm{border}}(\mathbf{x}_{\mathrm{ca}})
=\frac{d_{\mathrm{border}}(\mathbf{x}_{\mathrm{ca}})}{\max_{y\in E_{t}} d_{\mathrm{border}}(y)}.
\]

Additionally, we incorporate temporal cues: the tip positions form a continuous trajectory, and the tip position should be continuous while sampled points remain within the FoV. The temporal distance is

\begin{equation}
\small
\begin{aligned}
\tilde d_{\mathrm{temp}}^{\mathrm{bi}}(\mathbf{x}_{\mathrm{ca}})
&=\tfrac{1}{2}\Bigg(
\underbrace{\frac{\left\lVert \mathbf{x}_{\mathrm{ca}}-\tilde{\mathbf{p}}_{\mathrm{tip},\,t-1}\right\rVert_2}
{\max\limits_{y\in E_{t}}\left\lVert y-\tilde{\mathbf{p}}_{\mathrm{tip},\,t-1}\right\rVert_2}}_{\text{to previous-frame tip }(t{-}1)\ \text{normalized over candidates }E_{t}}
\\[-2pt]
&\qquad\quad+\;
\underbrace{\frac{\left\lVert \mathbf{x}_{\mathrm{ca}}-\tilde{\mathbf{p}}_{\mathrm{tip},\,t+1}\right\rVert_2}
{\max\limits_{y\in E_{t}}\left\lVert y-\tilde{\mathbf{p}}_{\mathrm{tip},\,t+1}\right\rVert_2}}_{\text{to next-frame tip }(t{+}1)\ \text{normalized over candidates }E_{t}}
\Bigg).
\end{aligned}
\label{eq:bi_temp_norm}
\end{equation}

The mixed score is given in~\eqref{equ:mix_loss_weak_supervised}, and the tip at frame \(t\) is obtained by the optimization in~\eqref{equ:optimization_weak_supervised}:
\begin{subequations}
\label{eq:bidirectional_tip_selection}
\begin{align}
s_t(\mathbf{x}_{\mathrm{ca}})
&=\alpha\,\tilde d_{\mathrm{border}}(\mathbf{x}_{\mathrm{ca}})
+\left(1-\alpha\right)\!\left[1-\tilde d_{\mathrm{temp}}^{\mathrm{bi}}(\mathbf{x}_{\mathrm{ca}})\right], \label{equ:mix_loss_weak_supervised}\\
\hat{\mathbf{p}}_{\mathrm{tip},\,t}
&=\arg\max_{\mathbf{x}_{\mathrm{ca}}\in E_{t}} s_t(\mathbf{x}_{\mathrm{ca}}). \label{equ:optimization_weak_supervised}
\end{align}
\end{subequations}
where $\alpha\in[0,1]$.

Before scoring, we suppress all possible tip positions (one per frame) within a small pixel threshold \(\tau\) of the boundary to encode the entry-point prior: \(E_{t} \leftarrow E_{t}\setminus\mathcal{B}_\tau\), where \(\mathcal{B}_\tau=\{\mathbf{x}_{\mathrm{ca}}\in E_{t}:\, d_{\mathrm{border}}(\mathbf{x}_{\mathrm{ca}})\le \tau\}\). At the first or last frame, or whenever a backward/forward pass is unavailable, \(\tilde d_{\mathrm{temp}}^{\mathrm{bi}}\) reduces to the available one-sided normalized distance. Applied over the sequence, this optimization can be viewed as an Iterated Conditional Modes (ICM)-style block coordinate-ascent (forward-backward fixed-point) procedure on a 1D chain~\cite{wright2015coordinate}.

The integrated pipeline operates autonomously by recording synchronized microscope video and robot motion data for at least $T$ seconds, automatically generating tip annotations throughout the sequence. Each detected tip coordinate $\hat{\mathbf{p}}_{\mathrm{tip}}^t$ generates a corresponding Gaussian heatmap to serve as the spatial supervision targets (pseudo-labels). The training loss can  be calculated through:
\begin{equation}
H_{i,j} \;=\; \exp\!\Bigl(-\tfrac{\|(i,j)-\hat{\mathbf{p}}_{\mathrm{tip}}^t\|^2}{2\sigma^2}\Bigr),
\end{equation}
where i,j denotes pixel coordinates in image. The variance parameter $\sigma^2$ controls the spatial spread of the probability distribution in pixel coordinates.

\subsection{Generation of Focal-Plane Distance via Kinematics}

At each anchor pixel $a_k$, the robot performs an up-down scan along the microscope’s optical axis (calibrated via the hand-eye calibration established in the warm-up) while keeping the tool tip in the FoV.
Given the automatically labeled tip position \(\hat{\mathbf{p}}_{\mathrm{tip},\,t}\) in frame \(I^t\), we extract an \(L\times L\) patch $P^t$ around it.

In many microscope setups, the optical path crosses multiple media (varying refractive indices), so absolute depth cannot be reliably inferred from single-frame defocus analysis. Instead, for each anchor \(a_k\), let \(\mathcal{T}_a\) denote the frame indices collected during its dedicated axial sweep (with \(\|\hat{\mathbf{p}}_{\mathrm{tip},\,t}-a_k\|_2\le \varepsilon\) by design). Let \(z_{t}\) be the robot-reported axial position at frame \(t\). We estimate the focal plane at anchor \(a_k\) by the maximal-sharpness frame~\cite{pertuz2013analysis}:
\begin{equation}
t_a^\star \;=\; \arg\max_{t\in\mathcal{T}_a}\ \frac{1}{|\Omega|}\sum_{(i,j)\in\Omega}\!\big(G_x^t(i,j)^2+G_y^t(i,j)^2\big),
\end{equation}
where \(\Omega\) indexes the patch, \((G_x^t,G_y^t)=\nabla P^t\), and \(\nabla\) is a Sobel operator. 
Instead of calibrating an absolute mapping from defocus to metric depth, we adopt a local, self-referenced approach. For each anchor, we identify the focal-plane depth $z^{\star}$ via a short axial scan and define it as the reference (z=0). Consequently, each crop in \(\mathcal{T}_a\) is assigned a relative label \(\Delta z_{t} := z_{t} - z^\star\), representing the signed stage displacement along the optical axis.

\subsection{Uncertainty-Aware Hand-Eye Calibration}\label{sec:markerless_handeye}

Leveraging the synchronized kinematic-visual data collected during the warm-up phase, we perform an offline, markerless calibration to establish the rotation $\mathbf{R} \in SO(3)$ between the robot base and the camera frame. This allows us to map visual error vectors into robot actuation commands.
Critically, we perform this calibration rather than relying on a one-time static calibration or a CAD model. This is because the micropipette is manually mounted and varies in length across experiments, rendering the CAD-derived tool center point (TCP) inaccurate (i.e., the tip-to-end-effector translation is unknown).
By recovering $\mathbf{R}$ from the motion data, we align the control frames without requiring prior knowledge of the variable tool geometry.

\paragraph{Assumptions}
Time-varying frame misalignment (latency jitter and sampling mismatch) and demonstration-induced visibility gaps (the tip intermittently leaves the FOV) still affect calibration. 
In practice, frames without tip detections are discarded and the corresponding robot poses are removed as well. Thus, the calibration operates on the co-visible subset. 
Within this subset, the remaining effect is time-varying correspondence uncertainty (e.g., $\pm$1-4 frame jitter, typical in software-synchronized data streams). Bidirectional Chamfer (Bi-Chamfer) aligns the two point sets without relying on fixed one-to-one pairing; the velocity-consistency term imposes a differential constraint that prevents erroneous correspondence at trajectory revisit points, where spatial proximity acts as a confounding factor for temporal alignment. The assumptions in the calibration module are:

\begin{enumerate}
\item \textbf{Approximate nominal intrinsics}: 
The image-plane scale $s$ and principal point $\mathbf c$ are set from nominal microscope/camera parameters (e.g., magnification and pixel pitch) rather than from a dedicated intrinsic calibration. 

\item \textbf{Weak-perspective validity}: Over the working depth range, the orthographic/weak-perspective model holds.
\item \textbf{Session-constant offsets}: The tool-tip  offset in end-effector frame is constant within a data-collection session, but it may change across re-mounts.
\item \textbf{Extrinsic rotation is fixed and estimated}: $\mathbf R\in\mathrm{SO}(3)$ is constant within a session and is the only reported hand-eye parameter; the translation part of hand-eye calibration acts as session-specific offsets that are accounted for.
\end{enumerate}

\paragraph{Bi-Chamfer Based Uncertainty-Aware Loss}
Given per-detection image-plane tip locations \(\hat{\mathbf{p}}_{\mathrm{tip},\,i}\in\mathbb{R}^2\) at time step i, and robot end-effector pose \((\mathbf{t}_{\mathrm{ee},t},\mathbf{R}_{\mathrm{ee},t})\) at frame \(t\), the session-constant tool-tip offset can be defined as \(\mathbf{r}_{\mathrm{tip}}\in\mathbb{R}^3\) in the end-effector frame. The offset expressed in the coordinate frame of the robot base is:
\begin{equation}
\mathbf{p}^{\mathrm{3D}}_i=\mathbf{R}_{\mathrm{ee},i}\,\mathbf{r}_{\mathrm{tip}}+\mathbf{t}_{\mathrm{ee},i}.
\label{eq:p3d_tip_re}
\end{equation}

Let \(\mathbf{p}^{\mathrm{cam}}_i\in\mathbb{R}^3\) be the tip in the camera frame, \(\mathbf{t}_{\mathrm{vec}}\in\mathbb{R}^3\) the translation part of the hand-eye calibration transform, and \(\mathbf{c}=[c_x,c_y]^{\top}\) the principal point (pixels) of the camera intrinsics. The projected image point \(\hat{\mathbf{p}}_i=[u_i\ v_i]^{\top}\in\mathbb{R}^2\) is:
\begin{subequations}
\begin{equation}
\mathbf{p}^{\mathrm{cam}}_{i}=\mathbf{R}\,\mathbf{p}^{\mathrm{3D}}_i+\mathbf{t}_{\mathrm{vec}},
\label{eq:weak_persp_re1}
\end{equation}
\begin{equation}
\hat{\mathbf{p}}_i = \mathbf{c} + 
\mathbf{L}_{\rm img}\mathbf{p}^{\mathrm{cam}}_{i},
\label{eq:weak_persp_re2}
\end{equation}
\end{subequations}
where the interaction matrix is denoted as \(\mathbf{L}_{\rm img}=\begin{bmatrix}s&0&0\\[2pt]0&s&0\end{bmatrix}\)  with session-constant scale \(s>0\).

Let $\mathcal{C}_{\mathrm{proj}}(\mathbf{R}) = \{ \hat{\mathbf{p}}_i(\mathbf{R}) \}_{i=1}^{N}$ [$\hat{\mathbf{p}}_i(\mathbf{R})$ is hence abbreviated to $\mathbf{p}_i$ for the sake of simplicity] denote the set of projected 2D tip positions, computed using the current extrinsic rotation estimate $\mathbf{R}$ via~\eqref{eq:weak_persp_re2}. Here, the index $i$ enumerates the discrete robot poses (time steps) selected for the calibration sequence ($N$ samples in total).
Let the observed subset be 
$\mathcal{C}_{\mathrm{2D}}=\{\mathbf{q}_j^{\text{img}}\}_{j=1}^{M}$ 
with $\mathbf{q}_j^{\text{img}} := \hat{\mathbf{p}}_{\mathrm{tip},\,j}$ the detected 2D tip keypoints at frames with valid detections. 
Because of asynchrony and filtering, $N$ and $M$ may differ.
We minimize the Bi-Chamfer distance:
\begin{equation}
\begin{aligned}
\mathcal{L}_{\mathrm{Chamfer}}
& = \
\underbrace{\frac{1}{2N}\sum_{i=1}^{N}\min_{1\le j\le M}\big\|\mathbf{p}_i-\mathbf{q}_j^{\text{img}}\big\|_2}_{\text{projected}\ \to\ \text{detected: avg.\ NN distance (norm.\ by $N$)}} + \\
\qquad &
\underbrace{\frac{1}{2M}\sum_{j=1}^{M}\min_{1\le i\le N}\big\|\mathbf{q}_j^{\text{img}}-\mathbf{p}_i\big\|_2}_{\text{detected}\ \to\ \text{projected: avg.\ NN distance (norm.\ by $M$)}}
.
\end{aligned}
\label{eq:chamfer_obj}
\end{equation}

\paragraph{Velocity-Consistency Regulation}

Under the weak-perspective (orthographic) model \eqref{eq:weak_persp_re2} 
for small increments of the tip and camera pose, we have:
\begin{equation}
\Delta\mathbf{p}\ \approx\ \mathbf{L}_{\rm img}\,\mathbf{R}\,\Delta\mathbf{p}^{\mathrm{3D}}\;-\;\big(\,\mathbf{L}_{\rm img}[\mathbf{p}^{\mathrm{cam}}]_{\times}\big)\,\delta\boldsymbol{\Phi},
\label{eq:fixedJ_analysis}
\end{equation}
where $\delta\boldsymbol{\Phi} \in\mathbb{R}^3$  denotes the small-angle rotation vector of the camera extrinsic expressed in the camera frame.
It represents a first-order rotational perturbation between adjacent frames due to misalignment.
Writing \(\mathbf{p}^{\mathrm{cam}}=(x,y,z)^\top\),
\(
\mathbf{L}_{\mathrm{rot}}(\mathbf{p}^{\mathrm{cam}})= -\,s
\begin{bmatrix}
0 & -z & \ \ y\\
z & \ \ 0 & -x
\end{bmatrix}, 
\)
we define the velocity-level loss as
\begin{equation}
\mathcal{L}_{\mathrm{vel}}
\;=\;
\rho_{\mathrm{huber}}\!\left[
\Delta\mathbf{p}
-\mathbf{L}_{\rm img}\,\mathbf{R}\,\Delta\mathbf{p}^{\mathrm{3D}}
-\mathbf{L}_{\mathrm{rot}}(\mathbf{p}^{\mathrm{cam}})\,\delta\boldsymbol{\Phi}
\right],
\end{equation}
where $\rho_{\mathrm{huber}}(\mathbf e)$ is the Huber penalty with scale $\kappa>0$.
We use the Huber loss (rather than pure $\ell_2$) to softly clip residuals induced by visibility gaps. Because frames without tip detections are discarded, the finite difference between the last visible frame and the next visible frame leaves a gap and produces an artificially large ``velocity”.  Such gap-bridging increments are not informative for calibration and would otherwise overweight the objective; the Huber penalty limits their influence while remaining quadratic for small, well-aligned increments.

The constrained calibration is:
\begin{subequations}
\begin{align}
\min_{\ \Theta}~ &
\mathcal{L}_{\mathrm{Chamfer}}(\mathbf{R}(\Theta))+\lambda_{\mathrm{vel}}\mathcal{L}_{\mathrm{vel}}(\mathbf{R}(\Theta))\\[-2pt]
\text{s.t.}~ &
\underline{\Theta}\ \le_{\text{elem.}}\ \Theta\ \le_{\text{elem.}}\ \overline{\Theta},
\end{align}
\end{subequations}
where $\Theta=(\theta_x,\theta_y,\theta_z)$ are intrinsic ZYX (yaw-pitch-roll) Euler angles in the camera frame and the inequalities are element-wise. 
Bounds are chosen to avoid the ZYX singularity and to keep the optimization in the small-angle regime used by the velocity linearization (e.g., $\theta_x,\theta_y\!\in[-\theta_\perp,\theta_\perp]$, $\theta_z\!\in[\theta_z^{\min},\theta_z^{\max}]$).
Scalar \(\lambda_{\mathrm{vel}}\ge 0\) balances alignment and velocity consistency. 
The velocity constraint anchors the local motion geometry to the robot-reported increments, improving identifiability of \(\mathbf{R}\) under desynchronization. Finally, $\underline{\Theta}$ and $\overline{\Theta}$ are the lower and upper bounds of rotation. 

\paragraph{Effect of Desynchronization}
Standard Euclidean matching assumes strict temporal correspondence ($\mathbf{q}_t \leftrightarrow \mathbf{p}_t$). However, a time offset $k \delta t$ introduces an along-trajectory bias that contaminates the hand-eye calibration loss, where $k$ denotes the number of sampling periods $\delta t$ that differ between the observation and the projection; it is an integer and can be negative. Expanding the difference between an observation $\mathbf{q}_t$ and a neighborhood projection $\mathbf{p}_{t+k}$ yields:
\begin{equation}
\underbrace{\mathbf{p}_{t+k}-\mathbf{q}_{t}^{\text{img}}}_{\text{mismatch residual}}
\ \approx\
\underbrace{\mathbf{p}_{t}-\mathbf{q}_{t}^{\text{img}}}_{\text{1-to-1 error}}
\;+\;
\underbrace{\mathbf{L}_{\rm img}\,\dot{\mathbf{X}}^{\mathrm{cam}}(t)\,k\delta t}_{\text{time-induced drift}}
\;+\; \text{noise}.
\label{eq:desync_expansion}
\end{equation}
In a strict 1-to-1 setup ($k{=}0$), the time-induced drift is treated as a geometric error, forcing the optimizer to incorrectly adjust $\mathbf{R}$ to compensate for time delay (generating incorrect gradients). 

By minimizing the Bi-Chamfer distance, we implicitly search for the optimal shift $k$ that minimizes the projection of the error along the trajectory tangent. This effectively separates the errors:
\begin{subequations}
\label{eq:err_decoupled}
\begin{align}
\mathbf{e}_{\mathrm{pos},i}
:= \min_{k} \|\mathbf{q}_i^{\text{img}} - \mathbf{p}_{i+k}\| 
\approx \| \underbrace{\mathbf{p}_{t}-\mathbf{q}_{t}^{\text{img}}}_{\text{hand-eye calibration loss}} \|,
\label{equ: pos_err_chamfer}
\end{align}
where the along-trajectory mismatch is absorbed by the nearest-neighbor association, leaving only the geometric registration residual (perpendicular to motion) to drive the calibration of $\mathbf{R}$.

To recover the kinematic constraints lost by relaxing temporal correspondence, we incorporate the velocity-consistency term:
\begin{equation}
\mathbf{e}_{\mathrm{vel},i}
:= \Delta\mathbf{q}_i^{\text{img}} - \Delta\mathbf{p}_{i}
\approx
\underbrace{\mathbf{L}_{\mathrm{rot}}(\Delta \mathbf{X}^{\mathrm{cam}}_{i})\,\delta\boldsymbol{\Phi}}_{\text{orientation mismatch}}
\;+\; \epsilon_{\mathrm{noise}}.
\label{equ: vel_err_refined}
\end{equation}
Unlike position, the velocity increment $\Delta(\cdot)$ cancels out constant time offsets while remaining highly sensitive to rotational misalignment $\delta\boldsymbol{\Phi}$. Consequently, the combined objective allows us to filter out desynchronization noise (via Chamfer) while robustly optimizing extrinsic rotation (via Velocity), preventing the solver from fitting to timing artifacts.
\end{subequations}

\begin{figure}[t]
    \centering
    \includegraphics[width=1.0\columnwidth]{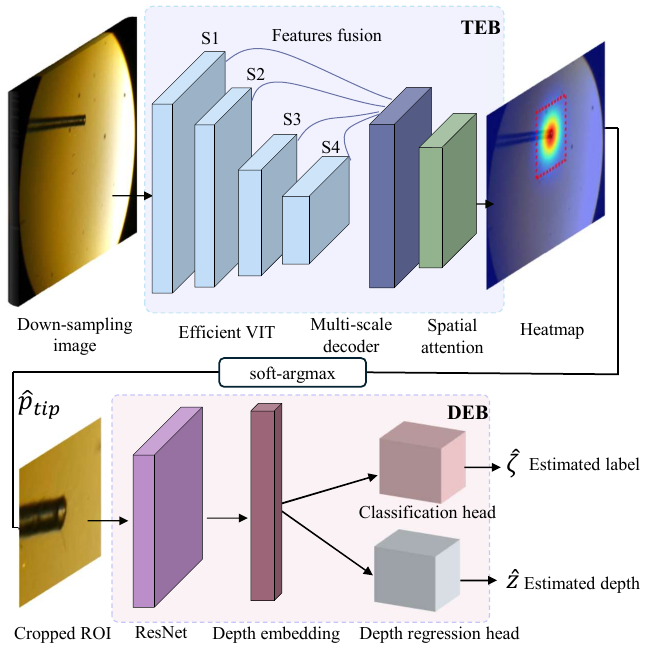}
    \caption{
    We deploy a two-stage detector for real-time 3D tip estimation. The tip estimation block (TEB) returns the keypoint observations and the depth estimation block (DEB) returns the normalized depth from tip to the focal plane and an auxiliary classification label.
    }
    \label{fig:TEBandDEB}
\end{figure}

\section{Real-time Tip Estimation Architecture}
\label{sec:RTTE}
Transitioning to online inference, this section details a 3D tracking architecture optimized for the low-latency demands of visual servoing. 
The proposed method integrates a 2D position block and a depth estimation block (Fig.~\ref{fig:TEBandDEB}) to provide real-time feedback ($\hat{\mathbf{p}}_{\mathrm{tip}}$). 
To ensure robustness, an estimator fuses these visual estimates with motion consistency prior.

\subsection{Real-time Tip Estimation Block}
The overall tracking architecture is structured into two cascaded stages: Stage 1 focuses on high-speed planar localization (Tip Estimation Block, TEB), while Stage 2 resolves the axial depth (Depth Estimation Block, DEB).
The TEB in stage~1 adopts a U-Net-like encoder-decoder architecture to preserve spatial feature resolution for target detection. However, unlike standard U-Nets that employ heavy convolutional encoders, we utilize EfficientViT as a lightweight backbone to strictly satisfy the latency constraints of the high-frequency control loop. A spatial attention block (shown in Fig.~\ref{fig:TEBandDEB}) is incorporated to enable efficient multi-crop feature fusion and rapid heatmap generation.

\subsubsection*{Resolution and Effective Precision}
For real-time inference, we employ a 10$\times$ down-sampling strategy. This design exploits the optical nature of the task: since the transparent tip is defined by high-contrast refractive edges rather than surface texture~\cite{xie2020segmenting}, the reduced resolution implicitly filters high-frequency optical noise (e.g., debris) while focusing the network on the dominant geometric structure~\cite{wang2020high}.

Critically, the spatial quantization loss from this 10$\times$ down-sampling is effectively recovered by the spatial soft-argmax operation, which has been proven to decouple localization precision from input resolution~\cite{sun2018integral}. 
This ensures that the system maintains high-fidelity tracking without the computational latency of processing full-resolution images.

\subsubsection*{Feature Fusion and Output}
We adopt a top-down, FPN-style fusion: encoder features at multiple strides are channel-aligned by $1\times1$ adapters, upsampled to a common resolution, and concatenated. A $3\times3$ fusion conv followed by a lightweight spatial-attention block reweights tip-salient regions before the heatmap head.
The fused feature is refined by stacked $3\times3$ Conv-BN-ReLU layers and upsampled stage-by-stage. The final pointwise convolution produces the predicted heatmap $\hat{\mathbf{H}} \in \mathbb{R}^{H'\times W'}$.
Subpixel localization is achieved via soft-argmax:
\begin{equation}
\mathbf{p}_{\mathrm{soft}} = \sum_{i,j}(i,j)\,\frac{\exp(\tau\,\hat H_{i,j})}{\sum_{u,v}\exp(\tau\,\hat H_{u,v})},
\end{equation}
where $\tau$ adjusts the distribution sharpness. 
The TEB is trained using tip-location pseudo-labels generated from sparsely prompted warm-up sequences.
The composite loss follows:
\begin{equation}
\mathcal{L}_{\mathrm{total}} = w_{\mathrm{MSE}}\,\mathcal{L}_{\mathrm{MSE}} + w_{\mathrm{Focal}}\,\mathcal{L}_{\mathrm{Focal}} + w_{\mathrm{bg}}\,\mathcal{L}_{\mathrm{bg}},
\end{equation}
where $\mathcal{L}_{\mathrm{MSE}}$ ensures reconstruction fidelity, $\mathcal{L}_{\mathrm{Focal}}$ handles class imbalance, and $\mathcal{L}_{\mathrm{bg}}$ suppresses background artifacts.

\subsection{Depth Estimation Block Structure and Learning}

To prevent overfitting to the limited warm-up samples, we attach a lightweight classification head that assigns the normalized axial offset
$\tilde{z} \triangleq \frac{z - z^\star}{s_z}$ from the focal plane into three classes:

\begin{equation}
\zeta(\tilde z)=
\begin{cases}
\texttt{below}, & \tilde z < -\delta,\\[2pt]
\texttt{near},  & -\delta \le \tilde z \le \delta,\\[2pt]
\texttt{above}, & \tilde z > \delta.
\end{cases}
\label{eq:depth-branching}
\end{equation}
where $z^\star$ is the focal-plane depth and $s_z$ is a normalization scale (e.g., training-range half-span). 
Thus, the auxiliary head classifies whether the tip is \emph{above} the focal plane, \emph{below} it, or \emph{nearly on} it (within $\delta$ of the normalized distance). 
This auxiliary signal improves representation compactness and provides a coarse, interpretable state for gain scheduling in the macro-micro controller.

We then implement a compact convolutional model, the Depth Estimation Block (DEB), for depth-aware representation learning under real-time constraints. The network supports both scalar depth regression and an auxiliary scene classification task within a low-dimensional embedding space.

DEB adopts a ResNet18 backbone configured for single-channel input and truncated at intermediate stages to obtain convolutional features from multi crops. Outputs from the last two stages are pooled and concatenated into a compact descriptor of dimension $\mathbf{x}_{\rm feature} \in C_{\text{feat}}$.
This descriptor is passed through an embedding head $C_{\text{feat}}\!\rightarrow\!\mathbb{R}^{d}$ (linear$\rightarrow$BN$\rightarrow$nonlinearity$\rightarrow$dropout). 
The embedding is then L2-normalized for scale invariance, leading to $\mathbf{\psi}\in\mathbb{R}^d$, with 
\(
\mathbf{\psi}=\dfrac{\phi(\mathbf{x_{\rm feature}})}{\|\phi(\mathbf{x_{\rm feature}})\|_2}.
\)
An embedding depth head maps the embedding $\psi$ to depth, and a classification head predicts a categorical state:
\begin{subequations}
\begin{equation}
\hat{z}=\underbrace{f_d(\mathbf{\psi})\in\mathbb{R}}_{\text{depth}}, \qquad
\end{equation}
\begin{equation}
\hat{\boldsymbol{\zeta}}=\underbrace{\mathrm{softmax}\!\big(f_c(\mathbf{\psi})\big)\in\mathbb{R}^K}_{\text{classification}},
\end{equation}
\end{subequations}
yielding class probabilities over the $K{=}3$ scene categories. The operator $\mathit{f}_d$, and $\mathit{f}_c$, represent  depth head, and classification head respectively. The $\hat{z}$ and $\hat{\zeta}$ represent estimated depth and the estimated label (class logits) in stage 2.

Under limited labels, depth regression and the auxiliary classifier can overfit to illumination/texture idiosyncrasies. To address this, we apply adversarial training to expose the model to small, worst-case image perturbations while enforcing prediction consistency.

Let $\mathcal{B}=\{(\mathbf{\psi}_i,z_i)\}_{i=1}^n$ be a mini-batch, where the index $i\in\{1,\dots,n\}$ enumerates samples. The normalized depth (ground truth) is $\tilde z_i$. For each anchor $i$, let $\mathcal{N}_k(i)$ be its $k$ nearest neighbors in the embedding space; define
$\mathbf{d}^{(\psi)}_i=\big[\|\mathbf{\psi}_i-\mathbf{\psi}_j\|_2\big]_{j\in\mathcal{N}_k(i)}$ and
$\mathbf{d}^{(z)}_i=\big[|\tilde z_i-\tilde z_j|\big]_{j\in\mathcal{N}_k(i)}$.
After per-vector normalization, we set
$\mathrm{dist}_{\mathbf{\psi}}\triangleq \mathrm{concat}_{i=1}^n \mathrm{norm}(\mathbf{d}^{(\psi)}_i)$ and
$\mathrm{dist}_{z}\triangleq \mathrm{concat}_{i=1}^n \mathrm{norm}(\mathbf{d}^{(z)}_i)$,
which stack the local pairwise distances in the embedding and depth, respectively. 
Given an input patch $P$ with embedding $\mathbf{\psi}$, depth $\hat z=f_d(\mathbf{\psi})$, and class logits $\hat{\boldsymbol{\zeta}}$, we minimize:
\begin{equation}
\begin{aligned}
\label{eq:loss_clean}
\mathcal{L}_{\text{clean}}(P)
&=
\underbrace{\lambda_d\,\|\hat z - \tilde z\|_2^2}_{\text{depth regression}}
+\underbrace{\lambda_s\,\big\|\mathrm{dist}_{\mathbf{\psi}}-\mathrm{dist}_{z}\big\|_2^2}_{\text{embedding-depth alignment}} \\
&\quad
+\underbrace{\lambda_v\!\left(-\mathrm{Var}(\mathbf{\psi})\right)}_{\text{variance}}
+\underbrace{\lambda_{\rm cls}\!\left(-\sum_i \hat{\boldsymbol{\zeta}}_i\log \hat{\boldsymbol{\zeta}}_i\right)}_{\text{classification entropy}},
\end{aligned}
\end{equation}
where the structural term aligns normalized $k$-NN pairwise distances in the embedding with absolute depth differences, encouraging locally smooth, control-friendly representations.

From a normalized input patch $P$, we construct an adversarial counterpart $P_{\mathrm{adv}}$ by Projected Gradient Descent (PGD) \cite{madry2017towards} within an $\ell_\infty$ ball of radius $\varepsilon$. $P_{\mathrm{adv}}$ can be updated iteratively, the t-th counterpart is denoted as  $P_{\mathrm{adv}}^{(t)}$, and the initial state $P_{\mathrm{adv}}^{(0)}=P$. The update law can be represented as:
\begin{equation}
\begin{aligned}
\label{eq:loss_adv}
P^{(t+1)}_{\mathrm{adv}}
=\Pi_{[x-\varepsilon,\,x+\varepsilon]}\!\Big(P^{(t)}_{\mathrm{adv}}
+\alpha\,\mathrm{sign}\!\big(\nabla_{x}\mathcal{L}_{\text{clean}}(P^{(t)}_{\mathrm{adv}})\big)\Big),
\end{aligned}
\end{equation}
where
$t$ is the PGD iteration index ($t{=}0,\dots,K{-}1$),
$\varepsilon>0$ bounds the perturbation in $\ell_\infty$ (max per-pixel change),
$\alpha>0$ is the PGD step size,
$\nabla_{x}\mathcal{L}_{\text{clean}}(\cdot)$ is the gradient of the clean loss \eqref{eq:loss_clean} w.r.t.\ the input,
$\mathrm{sign}(\cdot)$ applies element-wise sign to the gradient,
$\Pi_{[x-\varepsilon,\,x+\varepsilon]}(\cdot)$ projects back onto the $\ell_\infty$ ball around $x$ and clips to the valid pixel range (consistent with pre-processing).

We then optimize a convex combination of clean and adversarial losses with prediction-consistency regularization:
\begin{equation}
\begin{aligned}
\label{eq:loss_mix}
\mathcal{L}_{\text{mix}}
&=(1-\gamma)\,\mathcal{L}_{\text{clean}}(P)
+\lambda_{\text{cons}}^{z}\,\big\|\hat z(P_{\mathrm{adv}}^{(K)})-\hat z(P)\big\|_{1}\\
&+\gamma\,\mathcal{L}_{\text{clean}}(P_{\mathrm{adv}}^{(K)})+\underbrace{\lambda_{\text{cons}}^{\text{cls}}\,\mathrm{KL}\!\big(p_{\text{adv}}\parallel p_{\text{clean}}\big)}_{\text{KL divergence from $p_{\text{adv}}$ to $p_{\text{clean}}$}},
\end{aligned}
\end{equation}
where $\gamma\!\in\![0,1]$ controls the clean/adv mix, and $p_{\text{clean}},p_{\text{adv}}$ are softmax outputs on $x$ and $x_{\mathrm{adv}}$. Here, we use L1 loss as it grows linearly, so occasional large adversarial shifts don’t dominate the loss. The overall loss encourages depth and class predictions to remain stable under worst-case but realistic perturbations, improving depth-MAE stability and decision margins without increasing model size.

\subsection{Adapter-Style Tip Estimator for Closed-loop Control}

Our objective is to regulate the micropipette tip position $\mathbf{p}_{\mathrm{tip}, t}$ to reach a target location $\mathbf{p}^\star$ (updated via user prompts). To achieve this via closed-loop visual servoing, we design an estimator to track the system dynamics in the error space, providing real-time feedback on the deviation between the current tip position and the target.
\paragraph{State-Space Model}
Let the planar error state be $\mathbf{e}_t^{(x)}=[e_{x,t},\,e_{y,t},\,\dot e_{x,t},\,\dot e_{y,t}]^\top$ (representing position and velocity deviations relative to the current target $\mathbf{p}^\star$), and the depth error state be $\mathbf{e}_t^{(z)}=[e_{z,t},\,\dot e_{z,t}]^\top$.
The decoupled linear error dynamics are given by:
\begin{equation}
\begin{bmatrix}
\mathbf{e}^{(x)}_{t+1}\\[2pt]
\mathbf{e}^{(z)}_{t+1}
\end{bmatrix}
=
\begin{bmatrix}
\mathbf{F}_x & \mathbf{0}\\
\mathbf{0} & \mathbf{F}_z
\end{bmatrix}
\begin{bmatrix}
\mathbf{e}^{(x)}_{t}\\[2pt]
\mathbf{e}^{(z)}_{t}
\end{bmatrix}
+
\boldsymbol{\eta}_t
\label{eq:error_dynamics}
\end{equation}
Assuming a constant-velocity model with sampling interval $\Delta t$, the transition matrices are
$\mathbf{F}_x=\begin{bmatrix}\mathbf{I}_{2}&\Delta t\,\mathbf{I}_{2}\\ \mathbf{0}&\mathbf{I}_{2}\end{bmatrix}$ and
$\mathbf{F}_z=\begin{bmatrix}1&\Delta t\\ 0&1\end{bmatrix}$.
The process noise is $\boldsymbol{\eta}_t\sim\mathcal{N}(\mathbf{0},\,\mathrm{blkdiag}(\mathbf{Q}_x,\mathbf{Q}_z))$. Specifically, the planar process noise covariance $\mathbf{Q}_x(\Delta t)$ accounts for acceleration uncertainty $q_{0,x}$:
\begin{equation}
\mathbf{Q}_{x}(\Delta t) = q_{0,x}\begin{bmatrix}
\frac{\Delta t^3}{3}\mathbf{I}_2 & \frac{\Delta t^2}{2}\mathbf{I}_2\\[2pt]
\frac{\Delta t^2}{2}\mathbf{I}_2 & \Delta t\,\mathbf{I}_2
\end{bmatrix},
\end{equation}
and \begin{equation}
\mathbf{Q}_z(\Delta t) = q_{0,z}\begin{bmatrix}
\frac{\Delta t^3}{3}& \frac{\Delta t^2}{2}\\[2pt]
\frac{\Delta t^2}{2} & \Delta t\,
\end{bmatrix}\end{equation} accounts for acceleration uncertainty $q_{0,z}$.
For the measurement model, we unify the notation for $\chi\!\in\!\{x,z\}$:
\begin{equation}
\mathbf{y}_t^{(\chi)} \;=\;\mathbf{H}_\chi \mathbf{e}^{(\chi)}_{t} \;+\; \boldsymbol{\nu}_t^{(\chi)}.
\label{eq:unified_obs}
\end{equation}
For the planar case ($\chi{=}x$), the TEB detector returns the tip pixel
$\hat{\mathbf{p}}_{\mathrm{tip}, t}$. We define the observation as the pixel residual
$\mathbf{y}_t^{(x)} = \hat{\mathbf{p}}_{\mathrm{tip}, t} - \mathbf{p}^\star$. The observation
matrix is $\mathbf{H}_x = \begin{bmatrix} \mathbf{I}_2 & \mathbf{0} \end{bmatrix}$,
so that the first two components of $\mathbf{e}^{(x)}_t$ coincide with the
planar position error in pixel space. Note that while the filter operates in
pixel space, for control purposes we lift the estimated planar position
error to task space via
$\mathbf{e}^{(x)}_{\text{task},t}
 = (\mathbf{L}_{\mathrm{img}}\hat{\mathbf{R}})^{\dagger}
   \hat{\mathbf{e}}^{(x,\mathrm{pos})}_{t}$,
where $\hat{\mathbf{e}}^{(x,\mathrm{pos})}_{t}\in\mathbb{R}^2$ denotes the
position components of $\hat{\mathbf{e}}^{(x)}_{t}$. Similarly for depth
($\chi{=}z$), 
$\mathbf{H}_z = \begin{bmatrix} 1 & 0 \end{bmatrix}$.

\paragraph{Planar Update via Soft Gating}
For lateral motion, we employ a constant-velocity Kalman filter based on the error dynamics defined in \eqref{eq:error_dynamics}. We modulate its measurement update using a data-driven gate that determines how strongly the filter should trust each detection.
This reliability measure fuses two cues into a unified score $s_t$: \textbf{(i)} a nonconformity score derived from the Kalman innovation, which quantifies dynamic inconsistency compared to the calibration set~\cite{yang2023object}, and \textbf{(ii)} the detector's heatmap confidence $c_t$. Instead of hand-tuning parameters, we determine statistics on a held-out quantile-calibration set derived from the accumulated warm-up data.

Specifically, we compute the squared Mahalanobis distance of the innovation $\boldsymbol\varepsilon_t$:
\begin{equation}
    d_t^2 = \boldsymbol\varepsilon_t^\top \mathbf{S}_{t}^{-1} \boldsymbol\varepsilon_t,
    \label{equ:innov_mahal}
\end{equation}
where $\mathbf{S}_t$ is the innovation covariance.
On a calibration set $\mathcal D_{\mathrm{cal}}$, we compute the split-conformal radius
$q=\mathcal{Q}_{1-\alpha}(\{d_t^2:\ t\in\mathcal D_{\mathrm{cal}}\})$.
This radius normalizes the nonconformity, yielding a raw reliability score $s_t$:
\begin{equation}
    s_t = \min\bigl(c_t,\ \max(0, 1-d_t^2/q)\bigr) \in [0,1].
    \label{equ:s_fused}
\end{equation}
Finally, we map $s_t$ to the soft gating factor $g_t^{(x)}$ and the adaptive covariance $\mathcal R_t^{(x)}$:
\begin{subequations}
\begin{align}
    g_t^{(x)} &= \mathrm{clip}\big(\tfrac{s_t}{\tau_x},\,0,\,1\big), \label{equ:gate}\\
    \mathcal R_t^{(x)} &= \sigma_{x0}^2\,\exp(-\gamma_x\, s_t)\,\mathbf{I}_2, \label{equ:meas_cov}
\end{align}
\end{subequations}
where the threshold $\tau_x$ is chosen such that an $r_{\mathrm{full}}$ fraction of reliable frames receive full updates ($g_t^{(x)}{=}1$).

\paragraph{Depth Update via Multi-crop Fusion}
To mitigate estimation noise through spatial consensus, we compute a spatially averaged depth measurement. We extract crops $P_b$ at a set of fixed 2D offsets $\mathcal{B}$ around the detected tip $\hat{\mathbf{p}}_{\mathrm{tip}, t}$. Aggregating the depth error estimates $\{\hat e_{t,\mathbf b}| \hat e_{t,\mathbf b}=\hat{z}(P_b)\}_{\mathbf b\in\mathcal{B}}$ yields the scalar measurement and its associated uncertainty:
\begin{subequations}
\begin{align}
\mathbf{y}^{(z)}_t = \frac{1}{|\mathcal{B}|}\sum_{\mathbf b\in\mathcal{B}} \hat e_{t,\mathbf b}, \\
S^{(z)}_t = \mathrm{STD}\bigl(\{\hat e_{t,\mathbf b}:\mathbf b\in\mathcal{B}\}\bigr).
\end{align}
\end{subequations}
Here, the standard deviation $S^{(z)}_t$ serves as a proxy for estimation noise (disagreement among crops). We map this score to the soft gate and measurement covariance:
\begin{subequations}
\begin{align}
g^{(z)}_t &= \mathrm{clip}\Big(1 - \tfrac{S^{(z)}_t}{\tau_z},\,0,\,1\Big), \label{equ:gate_depth}\\
\mathcal{R}^{(z)}_t &= \sigma_{z0}^2\,\exp\big(\gamma_z\,S^{(z)}_t\big),
\end{align}
\end{subequations}
where $\tau_z$ is the uncertainty tolerance (the gate closes when deviation exceeds $\tau_z$), $\sigma_{z0}^2$ is the baseline noise variance, and $\gamma_z > 0$ scales the covariance growth as crop consistency degrades, effectively causing the filter to reject unreliable visual updates.

Dynamic boundary handling shifts the crop center when the tip is near image edges, ensuring the tool remains within the receptive field without padding artifacts.
To prevent oscillation at the setpoint, we enforce a strict stability criterion: motion halts only when three consecutive measurements fall within the target focal zone.

\paragraph{Recursive Filtering}
First, the time update propagates the error state and covariance forward using the constant-velocity dynamics from \eqref{eq:error_dynamics}. For both planar ($\chi{=}x$) and depth ($\chi{=}z$) branches, the a priori estimates are: 
\begin{subequations}
\begin{align}
\hat{\mathbf{e}}^{(\chi)}_{t|t-1} &= \mathbf{F}_\chi\,\hat{\mathbf{e}}^{(\chi)}_{t-1},\\
\mathbf{P}^{(\chi)}_{t|t-1} &= \mathbf{F}_\chi\,\mathbf{P}^{(\chi)}_{t-1}\mathbf{F}_\chi^\top + \mathbf{Q}_\chi.
\end{align}
\end{subequations}

The measurement update subsequently corrects the predicted estimates by incorporating the observation $\mathbf{y}^{(\chi)}_t$ in~\eqref{eq:unified_obs}.
Crucially, this step is modulated by the adaptive covariance $\mathcal{R}^{(\chi)}_t$ and scalar gate $g^{(\chi)}_t$ derived in \eqref{equ:meas_cov} and \eqref{equ:gate_depth}.
The innovation $\boldsymbol{\varepsilon}^{(\chi)}_t$, soft-gated gain $\mathbf{K}^{(\chi)}_t$, and the final \textit{a posteriori} estimates are computed as:
\begin{subequations}
\begin{align}
\boldsymbol{\varepsilon}^{(\chi)}_t &= \mathbf{y}^{(\chi)}_t - \mathbf{H}_\chi \hat{\mathbf{e}}^{(\chi)}_{t|t-1}, \\
\mathbf{S}^{(\chi)}_t &= \mathbf{H}_\chi \mathbf{P}^{(\chi)}_{t|t-1}\mathbf{H}_\chi^\top + \mathcal{R}^{(\chi)}_t, \\
\mathbf{K}^{(\chi)}_t &= g^{(\chi)}_t\,\mathbf{P}^{(\chi)}_{t|t-1}\mathbf{H}_\chi^\top \big(\mathbf{S}^{(\chi)}_t\big)^{-1}, \label{eq:gated_gain}\\
\hat{\mathbf{e}}^{(\chi)}_{t} &= \hat{\mathbf{e}}^{(\chi)}_{t|t-1} + \mathbf{K}^{(\chi)}_t \boldsymbol{\varepsilon}^{(\chi)}_t, \label{equ:error_feed} \\
\mathbf{P}^{(\chi)}_{t} &= \big(\mathbf{I}-\mathbf{K}^{(\chi)}_t\mathbf{H}_\chi\big)\mathbf{P}^{(\chi)}_{t|t-1}.\label{equ:error_P}
\end{align}
\end{subequations}
Crucially, the gate $g^{(\chi)}_t$ scales the Kalman gain in \eqref{eq:gated_gain}. When reliability drops (reflected by high measurement uncertainty in $\mathcal{R}_t^{(\chi)}$, $g^{(\chi)}_t \to 0$), forcing the filter to rely on the motion model. Conversely, high reliability ($g^{(\chi)}_t \to 1$) recovers the standard update\footnote{Proof can be found in Supplementary I Stability and Error Analysis of the Soft-Gated Estimator}. To preclude covariance divergence during prolonged signal loss, a watchdog timer resets the filter if low TEB heatmap persists beyond a safety horizon, ensuring bounded error covariance \cite{Sinopoli2004Kalman}.

\section{Unified Macro-Micro Control Framework}

To realize the proposed adaptive visual servoing, we adopt a unified macro-micro scheme resolving the scale discrepancy between force guidance and microscopic visual regulation.

\subsection{Macro Controller: Admittance-Guided QP}
\label{sec:macro_admittance_qp}

\paragraph{Admittance with Dead-zone}
Let $\mathbf{q}\!\in\!\mathbb{R}^{n_q}$ be joint values, $\mathbf{J}(\mathbf{q})$ the robot's geometric Jacobian (with translational block $\mathbf{J}_p$),
$\mathbf{p}(\mathbf{q}),\mathbf{R}(\mathbf{q})$ the end-effector pose, and
$\mathbf{w}_t=[\mathbf{f}_t^\top,\boldsymbol{\tau}_t^\top]^\top$ the input wrench measured at the robot tip.
We place a dead-zone $\mathrm{dz}_{\boldsymbol{\delta}}(\mathbf{\cdot})$ on the interaction force $\mathbf{f}_t$ to define the filtered force $\tilde{\mathbf{f}}_t$:
\begin{subequations}
\label{eq:admittance_dz}
\begin{align}
\tilde{\mathbf{f}}_t &= \mathrm{dz}_{\boldsymbol{\delta}}(\mathbf{f}_t), \\[2pt]
\big[\mathrm{dz}_{\boldsymbol{\delta}}(\mathbf{f})\big]_i&=\max\big(0, |f_i|-\delta_i\big)\,\mathrm{sgn}(f_i),
\end{align}
\end{subequations}
where $\delta_i$ prevents noise-induced motion.
With forward Euler integration (period $\Delta t$) and Cartesian-velocity saturation $\mathrm{sat}_{\mathbf{v}_{\max}}(\cdot)$, the desired admittance velocity and displacement are:
\begin{subequations}
\label{eq:adm_discrete}
\begin{align}
\mathbf{v}^{\mathrm{adm}}_{t+1} &= \mathbf{v}^{\mathrm{adm}}_{t}
 + \Delta t\,\mathbf{M}_d^{-1}\!\big(\tilde{\mathbf{f}}_t - \mathbf{B}_d\,\mathbf{v}^{\mathrm{adm}}_{t}\big),\\
\Delta\mathbf{p}^{\mathrm{adm}}_{t} &= \mathrm{sat}_{\mathbf{v}_{\max}}\!\big(\mathbf{v}^{\mathrm{adm}}_{t+1}\big)\,\Delta t,
\end{align}
\end{subequations}
where $\mathbf{M}_d,\mathbf{B}_d\succ 0$ denote the positive definite virtual inertia and damping gains.

\paragraph{QP-Based Joint Command}
Given the current state $\mathbf{q}_t$, target rotation matrix $\mathbf{R}_g$ (pre-defined to maintain a feasible tool orientation for workspace accessibility), and the admittance-derived expected increment $\Delta\mathbf{p}^{\mathrm{adm}}_{t}$, we compute the macro joint velocity $\dot{\mathbf{q}}^{\rm macro}$ via a quadratic program.
Here, $\mathbf{R}_g$ serves as a fixed orientation target to decouple rotational degrees of freedom from translational motion, mitigating Jacobian inaccuracies for translational control precision.
The optimization is formulated as:
\begin{subequations}
\begin{align}
\min_{\dot{\mathbf{q}}^{\rm macro}}\quad
&  \underbrace{\big\| \mathbf{J}_p(\mathbf{q}_t)\dot{\mathbf{q}}^{\rm macro}\,\Delta t - \Delta\mathbf{p}^{\mathrm{adm}}_{t}\big\|^2_{\mathbf{W}_p} \notag
}_{\text{position}}+ \\&\underbrace{\lambda_R\big\|\log\!\big(\mathbf{R}_{\mathbf{g}}^\top \mathbf{R}(\dot{\mathbf{q}}^{\rm macro}\Delta t+\mathbf{q}_t)\big)\big\|_2^2}_{\text{rotation}} + \notag \\
&  \lambda_{dq}\,\|\dot{\mathbf{q}}^{\rm macro}\|_2^2
\\
\text{s.t.}\quad
& \underline{\dot{\mathbf{q}}} \le_{\text{elem.}} \dot{\mathbf{q}}^{\rm macro} \le_{\text{elem.}} \overline{\dot{\mathbf{q}}}, \qquad\\
& \|\mathbf{J}(\mathbf{q}_t)\,\dot{\mathbf{q}}^{\rm macro}\|_2 \le v_{\max},
\end{align}
\label{eq:macro_qp_final}
\end{subequations}
where $\mathbf{W}_p\succeq 0$, $\lambda_R$, and $\lambda_{dq} > 0$ are weights.
Vectors $\underline{\dot{\mathbf{q}}}$ and $\overline{\dot{\mathbf{q}}}$ denote the element-wise lower and upper limits for joint velocities, while $v_{\max}$ constrains the absolute magnitude of the tool tip's velocity in Cartesian space.
This formulation ensures the macro motion is bounded and strictly compliant with the admittance reference.

\subsection{Micro Controller: Error-Driven Refinement }
\label{sec:micro_controller}
This subsection details the confidence-gated feedback law, the decoupling strategies for handling optical blur, and the final kinematic fusion with the macro layer\footnote{Proof can be found in Supplementary II Proof of Controller Stability}.
\paragraph{Controller Design with Visibility Guard}
We directly feed the estimated error states $\mathbf{e}_t^{(x)}$ and $\mathbf{e}_t^{(z)}$ from the filter of \eqref{equ:error_feed}  to a feedback law $\mathcal{K}$.
However, since the filter state propagates via prediction even when tip detections are absent or unreliable, blindly following the estimated error state can cause drift. To suppress autonomous commands based on unreliable visual estimates, we gate the control output using a visibility indicator, $\mathcal{I}_{\text{vis}}$, which adapts to the filter's initialization status.

During initialization (cold start), prediction errors in Kalman filter are undefined due to the lack of state history, so we rely on the raw detector confidence to validate tool tip presence: $\mathcal{I}_{\text{vis}} = \mathcal{I}(c_t > \tau_{\text{det}})$.
Once the filter initializes (warm start), we switch to the fused reliability score derived in \eqref{equ:s_fused}, which incorporates dynamic consistency to reject outliers: $\mathcal{I}_{\text{vis}} = \mathcal{I}(s_t > \tau_{\text{loss}})$.

The guarded micro-scale velocity is:
\begin{equation}
\mathbf{u}^{\mathrm{micro}}_t \;=\; \mathcal{I}_{\text{vis}} \cdot (\mathbf{u}^{\text{x}}_t, u^{\text{z}}_t)^\top \;=\; \mathcal{I}_{\text{vis}} \cdot \mathcal{K}\big(\hat{\mathbf{e}}^{(x)}_{t},\,\hat{\mathbf{e}}^{(z)}_{t}\big).
\end{equation}
When the tip is reliably tracked, $\mathcal{K}$ functions as a discrete-time LQR using the full state estimate (including velocity) to provide damping. If the tip is lost ($\mathcal{I}_{\text{vis}}=0$), the micro-controller is effectively disengaged ($\mathbf{u}^{\mathrm{micro}}_t=\mathbf{0}$), preventing motion based on open-loop filter predictions.
\paragraph{Auxiliary Virtual Constraints}

To enforce the lateral and depth control logic, we inject auxiliary constraints. While lateral visual servoing is globally controllable (i.e., robust to defocus), rapid lateral motion induces image blur that destabilizes real-time depth estimation. We mitigate this by decoupling the axes of motion. Let $\mathbf{e}^{(z)}_{\text{reg},t}$ denote the plane-keeping error, defined as the positional drift from the initial reference depth (locked at the motion onset) augmented with the instantaneous axial velocity.
We apply the following sequential logic:
\begin{itemize} 
\item \textit{Lateral-Dominant:}
During lateral alignment, we bypass the noisy real-time depth estimator. Instead, we enforce a virtual fixture that maintains the robot's axial position at the registered depth to prevent drift caused by motion blur:
\begin{align}
\mathbf{u}^{\text{x}}_t
&= \mathrm{sat}_{\textnormal{v}^{\text{x}}_{\max}}\!\big(K_{\text{x}}\,\hat{\mathbf{e}}^{(x)}_{t}\big), \label{eq:lat_dominant_lat}\\
u^{\text{z}}_t
&= \mathrm{sat}_{\textnormal{v}^{\text{z}}_{\max}}\!\big(K_{z}\,\mathbf{e}^{(z)}_{\text{reg},t}\big). \label{eq:lat_dominant_dep}
\end{align}

\item \textit{Depth-Dominant:}
Once lateral alignment converges or when specifically regulating focus, we suppress lateral motion and engage the real-time depth estimator:
\begin{align}
\mathbf{u}^{\text{x}}_t &= \mathbf{0}, \label{eq:depth_dominant_lat_zero}\\
u^{\text{z}}_t &= \mathrm{sat}_{v^{\text{z}}_{\max}}\!\big(K_{z}\,\hat e^{(z)}_{t}\big). \label{eq:depth_dominant_dep}
\end{align}
\end{itemize}

This asymmetric structure ensures robust lateral convergence without depth drift\footnote{This drift is caused by inaccurate inverse Jacobian matrix. When the inverse Jacobian is applied on virtual plane, decomposition of motion will lead to a deviation, in our cases, means there will be a residual motion out of plane.}, utilizing the global stability of the planar detector while mitigating the sensitivity of the depth estimator to motion blur.

\subsection{Fusion}
The micro twist $\mathbf{u}^{\mathrm{micro}}_t = [\mathbf{u}^{\text{x}}_t, u^{\text{z}}_t]$ (aligned with the robot task frame) is mapped to joint values via a damped pseudo-inverse Jacobian and fused with the macro command:
\begin{equation}
\dot{\mathbf{q}}_{\rm cmd} = \dot{\mathbf{q}}^{\rm macro}
 +\mathbf J_p^\dagger{}_\lambda(\mathbf q_t)\,\mathbf u^{\mathrm{micro}}_t,
\end{equation}
where $\mathbf J_p^\dagger{}_\lambda(\mathbf q_t)$ is the pseudo-inverse Jacobian derived from translational block $\mathbf{J}_p$.

This summation mechanism inherently handles the transition from the initial macro-only approach to the final fine-tuning. As detailed in Section~\ref{sec:micro_controller}, the micro command $\mathbf u^{\mathrm{micro}}_t$ is gated by the visibility indicator $\mathcal{I}_{\text{vis}}$. Consequently, when the tool is outside the FoV or the tracking is unreliable ($\mathcal{I}_{\text{vis}}=0$), the second term vanishes, resulting in pure admittance control. Once the tool is reliably detected, $\mathbf u^{\mathrm{micro}}_t$ becomes non-zero, seamlessly superimposing the autonomous precision alignment onto the user's manual guidance without requiring a discrete mode switch.

\section{Ablation Experiments: Calibration-aware Tip Estimation under a Markerless Pipeline}
\label{sec:results}
This section details the construction of a weakly supervised, marker-free visual-motor dataset across distinct microscope setups and introduces a robust markerless calibration pipeline that effectively mitigates temporal asynchrony to achieve superior alignment accuracy over standard baselines.

\subsection{Setup and Data Acquisition Protocol}
\begin{figure}[tb]
    \centering
    \includegraphics[width=1.0\linewidth]{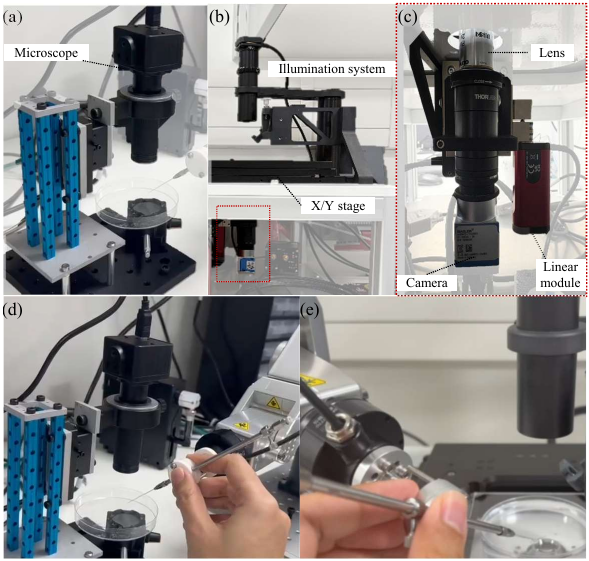}
    \caption{Dataset acquisition setup. (a) Upright microscope. (b-c) Inverted microscope with illumination control and a linear module for camera motion. (d-e) Steady-hand data collection under both setups.}
    \label{fig:image_setup}
\end{figure}

We built a marker-free visual-motor dataset by pairing promptable segmentation (SAM2) for clip-level mask propagation, followed by post-processing to obtain one temporally consistent tip endpoint (2D pixel) per frame. Timestamped frames were aligned with robot logs to pair pixels with robot poses. 
The microscope field of view corresponds to $s_{\mathrm{full}}=2.291\,\mu\mathrm{m}/\mathrm{px}$ at the $1920\times1440$ display and control-logging resolution. Images used for calibration and depth estimation were resized to $480\times360$, corresponding to $s_{\mathrm{depth}}=9.164\,\mu\mathrm{m}/\mathrm{px}$, while lateral tip inference was performed at $192\times144$, corresponding to $s_{\mathrm{lateral}}=22.91\,\mu\mathrm{m}/\mathrm{px}$. These scales assume that resizing preserves the same microscope field of view.
Compute was handled by a laptop (Intel i9-14900HX, NVIDIA RTX 4090).
We utilized two hardware configurations: an upright microscope and an inverted microscope (Fig.~\ref{fig:image_setup}).
For annotation, providing one prompt per clip ($\sim$500 frames) sufficed, reducing the burden to a one-click-per-clip workflow. 
Pseudo-label quality is subsequently verified through the calibration registration error: clips exceeding a data-driven threshold are flagged for targeted human review, replacing exhaustive frame-by-frame inspection with a principled, error-driven audit (Sec.~V-B-d, Appendix III).

\subsection{Dataset Composition}
To address the distinct challenges of planar localization and depth estimation, we constructed two specific sub-datasets.

\paragraph{Planar Tip Dataset ($\mathcal{D}_{\mathrm{tip}}$) for Detection and Calibration}
This dataset supports two downstream tasks: \textbf{(i)} Markerless Hand-Eye Calibration, aligning projected 3D tip trajectories with 2D visual tips; and \textbf{(ii)} Supervised Learning for the planar tip detector.
Data was collected via ``steady-hand" co-manipulation under both microscope setups. As the system transforms micromanipulation tasks into velocity commands rather than global positioning, the optimization only returns rotation (translation is not solved for). It is largely invariant to tool's image features, making the detector training \textit{setup-agnostic}. We pooled 8 trajectories collected under varying illumination conditions from both setups (approx. 45 mins total).
For evaluation, we reserved a held-out subset.
For the specific calibration task within this pipeline, we selected a single trajectory (the first 500 frames) from each setup to optimize the rotation matrix.

\paragraph{Depth Dataset ($\mathcal{D}_{\mathrm{depth}}$) for Axial Estimation}
Unlike planar features, defocus cues are highly sensitive to illumination and optics, making the depth estimator setup-dependent. Furthermore, the moving tip in $\mathcal{D}_{\mathrm{tip}}$ suffers from motion blur, which degrades depth supervision quality.
To mitigate this, we acquired a separate, clean depth sequence $\mathcal{D}_{\mathrm{depth}}$ ($\sim$22 mins) using the inverted microscope equipped with a motorized linear z-drive.
We executed wide lateral raster sweeps and axial motions; each frame was paired with the micron-level z-drive readout to serve as the relative depth ground truth. Note that while the z-drive provides precise relative motion, the absolute focal plane position is unknown to the robot; the agent learns to regulate depth relative to the sharpest focus (zero relative error) found during steady-hand sweeps.

\paragraph{Deployment and Inference}
The practical workflow is:
\textbf{(1)} Collect tool-only data for planar training ($\sim$45 min, valid across setups).
\textbf{(2)} Collect depth sampling ($\sim$22 min, per optic setup).
\textbf{(3)} For a specific session, calibrate the quantile-based soft gate using a brief sequence ($\sim$3 min, 1,909 samples).
Once trained, the system only requires the short markerless calibration step (using one 500-frame trajectory) if the robot base moves relative to the microscope.

\paragraph{Calibration as Pseudo-Label Audit}
\label{sec:calibration_audit}
The per-clip Bi-Chamfer registration error serves a dual role: beyond hand-eye estimation, it functions as an automatic quality indicator for pseudo-label correctness. If the detected 2D tip trajectory and the projected 3D robot motion are geometrically self-consistent, the pseudo-labels are reliable. 
Although the global calibration parameters are estimated from the same dataset, this evaluation is not circular: calibration minimises a global objective and absorbs only dataset-wide systematic bias, whilst clip-specific errors manifest as elevated local residuals that the global fit cannot mask.

We tested this mechanism on a representative set of 58 clips (27,203 frames, 25,073 valid detections, detection rate 92.2\%). Of 58 clips, 51 (87.9\%) were automatically approved; the 7 flagged clips were indicated to have genuine labeling issues upon qualitative inspection, including tip traversal of the air-liquid interface which obscures tip features (Appendix~III). 

For the dataset used in this study, the proposed workflow reduced the estimated annotation effort from 15.11\,h to 3.15\,h, corresponding to an estimated saving of $79.1\%$ relative to frame-by-frame annotation. These values are engineering estimates based on the measured workflow and an assumed frame-level annotation rate of approximately $2$\,s/frame for this task, rather than the outcome of a controlled annotation-time study. The workflow retains sparse prompting, automatic propagation, clip-level audit and correction, warm-up acquisition, axial depth sampling, and setup-specific calibration.

\subsection{Markerless Calibration}
In a GPU-driven visual-servo pipeline, the vision and motion processes can be temporally misaligned due to non-real-time operation system scheduling and imaging noise. While this is often negligible in conventional robotics, in micromanipulation it causes accuracy loss: the vision thread localizes the tip in frame \(t\), but the robot reports joint states at \(t+\Delta t\).

With the dataset $\mathcal{D_{\rm tip}}$, the optimizer minimizes a Bi-Chamfer loss between projected 3D points and 2D tips, plus a differential (velocity) consistency term with a fixed image Jacobian, and regularizes rotation on $\mathrm{SO}(3)$.
Decomposing residuals into position and velocity channels clarifies how small rotation errors and temporal misalignment manifest differently; importantly, distributional set alignment via Chamfer naturally absorbs small timestamp offsets that would otherwise break one-to-one correspondences.

\paragraph*{Baselines}
We compared against established approaches:
\textbf{PnP registration~\cite{barath2020magsac++}}; i.e. perspective-$n$-point optimization using manually selected correspondences.
\textbf{Jacobian regression~\cite{ZHU2021103798}}; i.e. regression of a constant image Jacobian from paired velocity deltas.
\textbf{Euclidean distance}; i.e. direct point-to-point matching serving as a strict correspondence baseline.

\begin{table}[t]
\centering
\caption{Hand-eye calibration accuracy across microscope configurations (mean$\pm$std).}
\label{tab:calibration_results_compact}
\resizebox{\columnwidth}{!}{%
\begin{tabular}{|l|c|c|c|c|}
\hline
\multirow{2}{*}{\textbf{Method}} & \multicolumn{2}{c|}{\textbf{Inverted Setup} (500 samples)} & \multicolumn{2}{c|}{\textbf{Upright Setup} (500 samples)} \\
\cline{2-5}
& \textbf{Reproj. [px]} & \textbf{Velocity [px/f]} & \textbf{Reproj. [px]} & \textbf{Velocity [px/f]} \\
\hline
PnP registration & 59.23$\pm$30.32 & 3.54$\pm$2.46 & 18.65$\pm$11.55 & 6.94$\pm$5.60 \\
\hline
Jacobian regression & 156.61$\pm$61.36 & 6.35$\pm$4.17 & 303.27$\pm$66.12 & \textbf{4.60$\pm$6.12} \\
\hline
Euclidean distance & 30.48$\pm$12.87 & 3.24$\pm$2.29 & 9.52$\pm$11.47 & 4.69$\pm$6.46 \\
\hline
\textbf{Proposed Method} & \textbf{4.86$\pm$2.72} & \textbf{2.91$\pm$1.96} & \textbf{8.41$\pm$11.83} & 4.81$\pm$6.66 \\
\hline
\end{tabular}%
}
\end{table}

\begin{table}[tb]
\centering
\caption{Asynchrony-induced error (Reproj$-$Chamfer) and pose error (Bi-Chamfer, symmetric mean), computed from the trimmed statistics (top 10\% removed). Euler angles are the optimized extrinsic rotations $(r_x, r_y, r_z)$ in radians.}
\label{tab:async_pose_errors}
\resizebox{\columnwidth}{!}{%
\begin{tabular}{|l|c|c|c|c|}
\hline
\textbf{Setup}  & \textbf{Reproj Mean [px]} & \textbf{Chamfer (Sym.) Mean [px]} & \textbf{Asynchrony Error [px]} \\
\hline
Inverted & 4.8692 & 2.2940 & \textbf{2.5752} \\
\hline
Upright  & 8.4175 & 1.4665 & \textbf{6.9510} \\
\hline
\end{tabular}%
}
\end{table}

We initialized an optimizer via the SVD projection to $\mathrm{SO}(3)$ and refined with a CasADi/IPOPT solver~\cite{Andersson2018}.
As shown in Table~\ref{tab:calibration_results_compact}, our Chamfer-based calibration achieves the lowest errors across both microscope geometries: $4.86\pm2.72$ px (inverted) and $8.41\pm11.83$ px (upright).

Table~\ref{tab:async_pose_errors} isolates two effects conflated in plain reprojection error. First, the asynchrony-induced error (Reproj$-$Chamfer) is substantially larger on the upright microscope (6.95 px) than on the inverted set-up (2.58 px), indicating that most of the residual on the upright configuration arises from timestamp misalignment rather than an incorrect extrinsic. Second, the pose error measured by the symmetric Bi-Chamfer mean remains small on both set-ups (2.29 px inverted; 1.47 px upright), consistent with the low spatial bias after calibration.

\begin{table*}[tb]
\centering
\caption{Unified comparison of (A) tool tip detection, (B) depth estimation and classification, and (C) adversarial robustness.}
\label{tab:unified_results}
\scriptsize
\renewcommand{\arraystretch}{1.15}
\begin{tabular}{lccccccc}
\toprule
\multicolumn{8}{c}{\textbf{(A) Tool Tip Detection at $\tau_{\mathrm{det}}=0.35$ (Testing samples 8{,}604 @ 192$\times$144)}} \\
\midrule
Model & Median Err. & FPR (\%) & FNR (\%) & Conf. (mean) & PCK@6 (\%) & PCK@10 (\%) & PCK@15 (\%) \\
\midrule
EfficientViT (Orig.)             & 2.83 & 3.70 & 10.86 & 0.478 & 91.61 & 93.90 & 94.60 \\
MobileVIT (Spatial Attn.)        & 2.24 & 6.94 & 6.92  & 0.619 & 94.42 & 95.07 & 95.31 \\
EfficientViT (Spatial Attn.)     & \textbf{1.41} & 7.03 & \textbf{6.50} & \textbf{0.665} & \textbf{95.08} & \textbf{95.95} & \textbf{96.22} \\
MobileNetV4                      & 2.24 & 8.80 & 7.22  & 0.565 & 94.01 & 94.95 & 95.07 \\
EfficientNet                     & 2.36 & 11.57 & 8.63 & 0.553 & 93.72 & 94.48 & 94.72 \\
\midrule
\multicolumn{8}{c}{\textbf{(B) Depth Estimation and Classification (Samples: 2{,}503)}} \\
\midrule
Model (dim) & MSE & R$^2$ & Cls. Acc. & F1 (macro) & F1 (C0) & F1 (C1) & F1 (C2) \\
\midrule
LDSENet (32D)             & 0.167 & 0.813 & 0.771 & 0.731 & 0.861 & 0.501 & 0.831 \\
ResNet-based (16D)        & \textbf{0.156} & \textbf{0.825} & \textbf{0.793} & \textbf{0.753} & 0.855 & 0.543 & 0.861 \\
FocDepthFormerNet (32D)   & 0.178 & 0.801 & 0.769 & 0.728 & 0.836 & 0.504 & 0.845 \\
\midrule
\multicolumn{8}{c}{\textbf{(C) Adversarial Robustness \& Tracking (Samples: 8{,}604 Planar / 2{,}503 Depth)}} \\
\midrule
Tracker & MeanErr (px) & Orig PCK (\%) & Adv PCK (\%) & 95\% Err (control px) & DepthErr (mm) & Depth95\% (mm) & Time (ms) \\
\midrule
Hungarian KF      & 6.43 & 96.2 & 78.4 & 8.94 & 0.217 & 0.382 & 5.0 \\
SiamRPN           & 5.32 & \textbf{99.2} & 88.8 & 10.0 & 0.287 & 0.698 & 8.1 \\
SoftgateKF (Ours) & \textbf{5.31} & 97.7 & \textbf{94.2} & \textbf{8.25} & \textbf{0.214} & \textbf{0.356} & 8.3 \\
\bottomrule
\end{tabular}
\end{table*}
Although lateral network inference uses $192\times144$ inputs, all planar tracking errors in Table~\ref{tab:unified_results} are reported after mapping the predictions to the $480\times360$ control coordinate system.

\subsection{Detection, Depth, and Tracking Evaluation}

We evaluated the vision pipeline on a diverse dataset collected from two distinct microscope setups (see Fig.~\ref{fig:image_setup}) under heterogeneous illumination.
The primary dataset consists of 8 lateral sweeping trajectories ($\approx$45\,min, 27{,}203 frames at 10\,Hz), split 4:1 for training/validation. 
A separate test set (8{,}604 frames) includes challenging backgrounds with distractors (bubbles, particles).
For depth evaluation, we utilized the specific subset of 2{,}503 frames with ground-truth z-drive readings.
The lateral models used $192\times144$ network inputs. Their predicted tip coordinates were mapped to the $480\times360$ control coordinate system before tracking and control, corresponding to a physical scale of $s_{\mathrm{ctrl}}=9.164\,\mu\mathrm{m}/\mathrm{px}$. Models within each comparison used identical preprocessing and training settings.
Table~\ref{tab:unified_results} presents a unified comparison across three tasks: Planar Detection, Depth Estimation, and Adversarial Tracking.

\subsection{Planar Tip Detection}
We calculated standard metrics: Median Localization Error, False Positive/Negative Rates (FPR/FNR) at $\tau_{\mathrm{det}}=0.35$, and Percentage of Correct Keypoints (PCK) at thresholds $\delta \in \{6, 10, 15\}$ pixels.
As shown in Table~\ref{tab:unified_results}~(A), the \textbf{EfficientViT with Spatial Attention} achieves the best overall performance.
It attains the lowest median error (1.41\,px) and highest PCK scores (\textbf{96.22}\% @15\,px), outperforming the vanilla EfficientViT (PCK 94.60\%) and EfficientNet.
Crucially, spatial attention balances errors well (FPR 7.03\%, FNR 6.50\%), whereas EfficientNet suffers from high false positives (11.57\%) due to distractors, and standard EfficientViT tends to miss tips (FNR 10.86\%).

Qualitative analysis of the residual failure cases reveals two primary physical challenges associated with the microscopic environment.
First, when the tool tip crosses the air-liquid interface, surface tension induces a meniscus that creates a localized dark-field effect, obscuring the tip features.
Second, dynamic occlusions occur when floating debris (e.g., bubbles or particles) drift between the backlight source and the tool, blocking the optical path and causing momentary target loss.

\subsubsection{Depth Estimation}
Estimating the axial position of a transparent micropipette from monocular defocus cues presents unique challenges. Unlike standard depth-from-defocus tasks, this scenario involves a transparent object traversing an air-liquid interface under variable illumination, resulting in complex optical aberrations and a scarcity of representative training samples.
Consequently, effective solutions must rely on backbones capable of rapid fitting on limited datasets while satisfying strict real-time inference constraints.
Guided by this, we benchmark against two architectures optimized for efficient feature extraction in specialized domains:
\textbf{LDSENet}~\cite{JIN2025103216}, a lightweight dual-stream encoder designed for joint depth-semantic learning, and
\textbf{FocDepthFormerNet}~\cite{kang2024focdepthformer}, which utilizes transformer-based attention to capture long-range dependencies in focus cues.

We compared these against our compact dual-task model (16D ResNet-based) on a subset of 2{,}503 samples.
Depth targets are defined as relative distance to the focal plane, classified into three zones (C0: above, C1: near-focus, C2: below).
Evaluation metrics include regression accuracy (MSE, $R^2$), classification F1-scores.

Table~\ref{tab:unified_results}~(B) shows that our \textbf{ResNet-based (16D)} model achieves the best regression performance ($R^2=0.825$, MSE=0.156) and overall classification accuracy (0.793), outperforming the heavier 32D baselines.
This suggests that for this specific, data-limited physical task, a concise feature representation generalizes better than deeper, more complex architectures.
Per-class analysis reveals high precision for far-focus zones C0/C2 ($\approx$0.97) but expected ambiguity in the near-focus zone C1 (F1 $\approx$0.54), a limitation shared by all methods due to the physical depth-of-field constraints (i.e., the ``dead zone" where blur is indistinguishable). The regression and classification results further indicate that the learned representation preserves depth-dependent visual information across the evaluated focal range.

\subsubsection{Robust Tracking under Perturbation}
Finally, we evaluated the system's resilience to adversarial perturbations. We simulated optical disturbances (glints, occlusions) by injecting randomized Gaussian noise into the heatmaps of the test set.
We compared our \textbf{SoftgateKF} (Soft-Gated Kalman Filter in \ref{sec:RTTE} C) against a standard Hungarian Kalman Filter and SiamRPN tracker using identical calibration parameters.
The soft gate is calibrated on a held-out set to achieve 95\% empirical coverage ($\tau_x \approx 6.30$).

Results in Table~\ref{tab:unified_results}~(C) demonstrate the robustness of our conformal gating approach.
While SiamRPN excels on clean data (PCK 99.2\%), its performance drops significantly under attack (Adv PCK 88.8\%).
In contrast, \textbf{SoftgateKF} maintains superior robustness (Adv PCK \textbf{94.2}\%) and achieves the lowest empirical 95th-percentile tracking error (95th-percentile error: \textbf{8.25}\,px vs.\ 10.0\,px for SiamRPN).
The Hungarian KF, while fastest (5.0\,ms), fails to reject outliers effectively (Adv PCK 78.4\%).
Crucially, robust planar tracking directly benefits depth estimation: SoftgateKF yields the lowest depth error (Mean \textbf{0.214}\,mm), as it correctly rejects perturbed frames that would otherwise lead to sampling wrong depth patches.
The total inference time (8.3\,ms) fits comfortably within the 30\,Hz control cycle (aligning to the sampling rate of camera).

\section{Experimental Validation on Real-time Robotic Controller}
We evaluated the proposed framework on a physical micromanipulation platform, quantifying accuracy in both lateral and axial dimensions.

\subsection{System Configuration and Workflow}
\begin{figure}[t]
    \centering
    \includegraphics[width=1.0\linewidth]{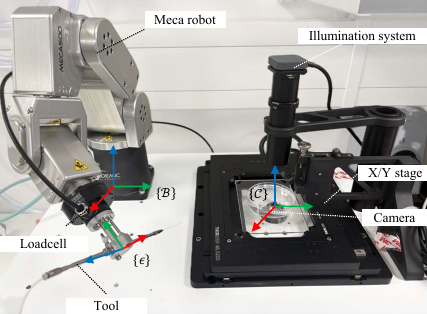}
    \caption{Robotic micromanipulation setup. A Meca500 arm with a loadcell operates under an inverted microscope. The system integrates manual macro-guidance and automated visual servoing.}
    \label{fig:setup}
\end{figure}

The user interface renders the microscope feed at $1920{\times}1440$ for operator viewing, while network inference operates on downsampled streams ($192{\times}144$ for lateral, $480{\times}360$ for depth) to ensure real-time performance.
A unified control panel integrates the robot, syringe pump, and linear stage, facilitating seamless macro-micro transitions.
The experimental setup comprises a Meca500 six-DOF robot arm equipped with a needle holder, operating under an inverted microscope (with a $2\times$ magnification lens). 
To enable macro-scale admittance control, a force sensor (MINIONE PRO, BOTA Systems) is mounted at the end-effector. 
Fluid actuation is provided by a syringe pump (HAMILTON PSD4) integrated into the system. 
A 250\,$\mu$L syringe (1750.5 BFP) is fitted to the pump, with tubing connecting the pump valve to the tool holder (HI-9, Narishige). 
The tool tip (275\,$\mu$m Stripper, CooperSurgical) is a consumable component and is replaced before each experiment. 
Coordinate frames are defined as shown in Fig.~\ref{fig:setup}: robot base $\{\mathcal{B}\}$, end-effector $\{\mathcal{E}\}$, camera optical frame $\{\mathcal{C}\}$, and pixel frame $\{\mathcal{I}\}$.
Intrinsic parameters are pre-calibrated, and the hand-eye transformation is determined via the markerless calibration pipeline described in Sec.~\ref{sec:results} C.

\begin{figure*}[t]
    \centering
    \includegraphics[width=1.0\linewidth]{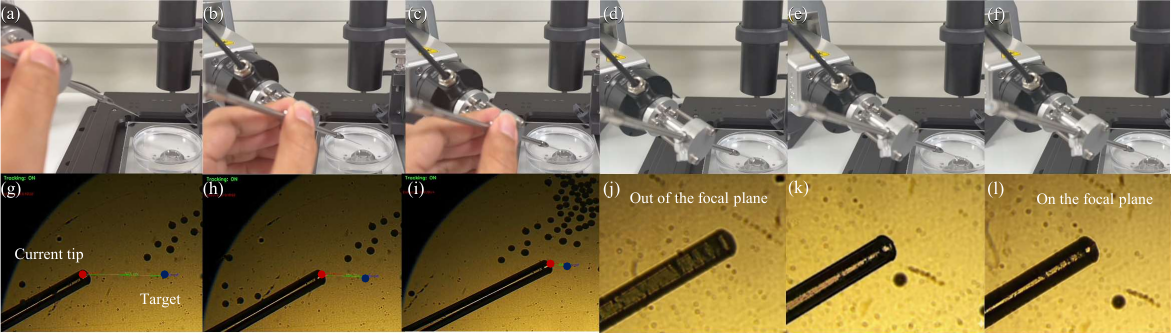}
    \caption{Shared control workflow: (a-c) Manual macro-guidance brings the tool into the FOV; (d-f) Automated visual servoing refines position; (g-i) Microscope view during lateral alignment; (j-l) View during depth regulation.}
    \label{fig:workingpipeline}
\end{figure*}

The macro-to-micro workflow (Fig.~\ref{fig:workingpipeline}) proceeds in three phases:
(1) \textbf{Macro Approach:} The operator physically guides the tool into the microscope's FoV.
(2) \textbf{Visual Servoing:} Once visible, the system engages the complementary controller. The operator clicks a target on the screen, and the robot autonomously aligns the tip laterally while regulating depth.
(3) \textbf{Actuation:} The user commands specific tasks (e.g., aspiration/dispense) via the interface.

\subsection{Depth Control Performance}
\label{exp:depth_independent}

Since axial position is inferred rather than directly measured, we first evaluate the depth regulation accuracy in isolation.
We compared our learning-based estimator against a standard ``Focal Depth Control'' baseline, which employs an iterative hill-climbing search to maximize image sharpness gradients.
Using a motorized linear stage to shift the focal plane by controlled amounts (0--4\,mm), we task both controllers to return the tip to the focal plane.
For each position, we conducted 7 repeated trials with randomized initial lateral positions, reporting the mean and standard deviation.

To ensure robust inference, our method extracts $80{\times}80$ pixel patches centered at the detected tip. We collected samples of the depth filter by aggregating predictions from four crop offsets ($\pm10$\,px from the center) and the crop without offset.

\begin{figure}[tb]
    \centering
    \includegraphics[width=1.0\linewidth]{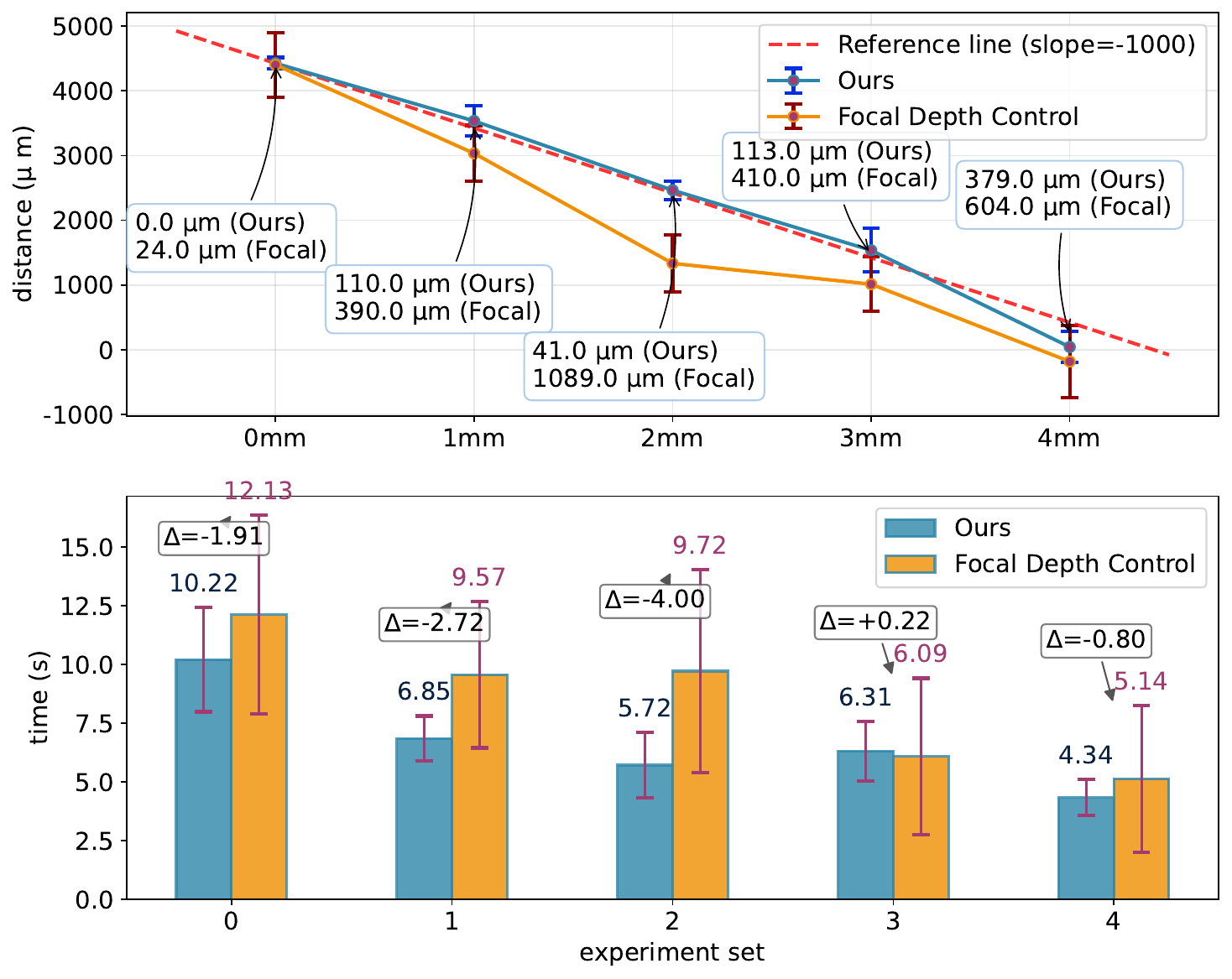}
    \caption{Depth regulation performance. Top: Final positioning error vs. initial displacement (Mean $\pm$ Std over 7 trials). Bottom: Convergence time. Our method (blue) shows a linear descent trend, indicating robust coverage over a larger depth range compared to the non-linear baseline (orange).}
    \label{fig:depthpipeline}
\end{figure}

The baseline utilizes a similar stopping logic based on the regional average defocus score. Crucially, to ensure a fair comparison under uneven illumination, we pre-calibrated the baseline's target defocus value for each spatial region, mitigating instability caused by local brightness variations. Results are summarized in Fig.~\ref{fig:depthpipeline}.

\textbf{Accuracy \& Range:} Our method demonstrates superior precision across all displacements. At 2\,mm offset, our mean error is $41\,\mu\text{m}$ versus $1089\,\mu\text{m}$ for the baseline; even at the challenging 4\,mm offset, our error ($379\,\mu\text{m}$) remains substantially lower than the baseline ($604\,\mu\text{m}$).
The trend analysis in Fig.~\ref{fig:depthpipeline} (Top) reveals that our method follows a quasi-linear descent, indicating its capacity to regress depth over a wider operational range. In contrast, the baseline relies on non-linear sharpness gradients, which require complex calibration and exhibit poor adaptability to large environmental changes.

\textbf{Speed:} Our method converges significantly faster (e.g., 5.72\,s vs 9.72\,s at 2\,mm offset) because it directly regresses the target step size, avoiding the latency of iterative gradient search. The baseline suffers from local optima in textureless regions, whereas our data-driven approach infers absolute distance robustly.

We observe a gradual degradation in estimation accuracy beyond the 4\,mm range. This behavior is expected, as our training dataset and calibration were primarily acquired at the nominal focal plane ($z{=}0$\,mm). 
However, in realistic micromanipulation scenarios, the operational focal plane typically remains within a bounded range. Consequently, data acquisition can be streamlined by focusing on the most frequent operational plane. 
Our results demonstrate that the learned model generalizes well to moderate focal shifts, successfully handling the associated variations in illumination and visual features, thereby validating its robustness. 
While aggregating models trained across multiple focal planes could theoretically extend the working range, it would necessitate a multiplicative increase in sampling time and data volume, which exceeds the practical requirements of this specific application.

Furthermore, achieving precise alignment between the tool tip and the target focal plane implies that the system has functionally bypassed the dominant task-critical geometric uncertainty in this setup, namely the unknown tip kinematics introduced by tool re-mounting. Rather than explicitly reconstructing the full chain \textit{Robot} $\leftrightarrow$ \textit{Unknown End-Effector} $\leftrightarrow$ \textit{Tip} $\leftrightarrow$ \textit{Focal Plane} $\leftrightarrow$ \textit{Z-axis Module}, the framework closes the loop through relative visual feedback along the optical axis.
This closes the task-critical tip-regulation loop through relative visual feedback without requiring explicit identification of the variable tool length or the absolute focal-plane position, thereby addressing the dominant tip-installation uncertainty for the demonstrated task.

\subsection{3D Tip Localization and Control}

\begin{figure}[t]
    \centering
    \includegraphics[width=1.0\linewidth]{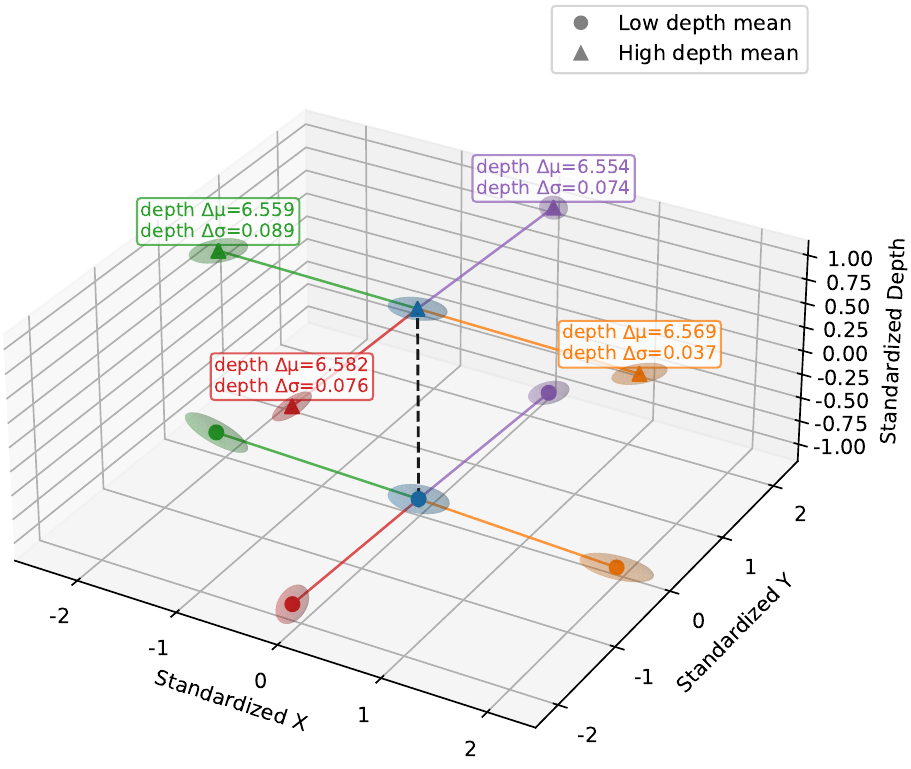}
    \caption{3D validation routine visualized in normalized task space. The lateral axis is normalized to $[-2, 2]$ and the depth axis to $[0, 1]$. Clusters represent the steady-state landing positions at Center (Blue) and Edge (Orange) targets. The reference depth ($0$\,mm, normalized to $0$) is defined by the center of the bottom-most blue cluster (the initial low-depth starting point).}
    \label{fig:3dview}
\end{figure}

To validate the full 3D complementary controller, we executed a multi-stage ``Center-Edge-Center'' trajectory pattern (Fig.~\ref{fig:3dview}), repeated three times to characterize observed variation.
The protocol integrates human-guided lateral targets with automated depth regulation in a specific sequence to ensure safety and coverage:
\begin{enumerate}
    \item \textbf{Low-Depth Lateral Phase:} Starting from an arbitrary depth below the focal plane, the operator guides the tip via mouse clicks from the Center to each image boundary (Top, Bottom, Left, Right) and back. This tests lateral servoing under significant defocus.
    \item \textbf{Center-Based Depth Regulation:} To mitigate the risk of tip loss at image boundaries during axial motion, the ``tip-back-to-focal-plane'' maneuver is executed \textit{only} when the tip is at the Center position.
    \item \textbf{High-Depth Lateral Phase:} Once at the focal plane, the \textit{virtual axial constraint} is engaged. The operator repeats the Center-Edge-Center pattern. This tests lateral precision while the controller suppresses axial drift.
\end{enumerate}
A trial was defined as one Center--Edge--Center lateral excursion at either the low-depth or high-depth condition. The four edge directions were evaluated under both depth conditions over three repetitions, yielding $4\times2\times3=24$ trials. 
A trial was counted as completed if the commanded servo action terminated in a stable landing point (the points entering the dispersion statistics below) without a tip-loss event (visibility guard opening, $\mathcal{I}_{\text{vis}}=0$, Sec.~\ref{sec:micro_controller}), without operator re-acquisition, and without a safety-triggered stop; $24/24$ trials met this definition. 

\paragraph{Axial Accuracy Analysis}
Fig.~\ref{fig:3dview} visualizes the spatial distribution of the landing points. Each cluster represents the steady-state position where the system stabilized after a user command.
The vertical axis represents depth, normalized to $[0, 1]$, where the origin corresponds to the mean depth of the initial low-depth central cluster (bottom blue cluster).
We assess depth consistency by measuring the dispersion $\sigma_z$ of the high-depth clusters relative to the target focal plane.
The worst-case standard deviation was $\sigma_{\max} = 89\,\mu\text{m}$.

Across the evaluated large in-plane motions, the axial estimates exhibited a systematic bias of $\bar{e}_z\approx113\,\mu$m, while the largest observed standard deviation was $\sigma_{\max}=89\,\mu$m. Combining the magnitude of this bias with twice the largest observed dispersion gives the following conservative engineering-level estimate:
\[
e_{z,\mathrm{eng}}
=2\sigma_{\max}+\bar{e}_z
\approx178+113=291\,\mu\mathrm{m}.
\]

This quantity summarizes the observed bias and dispersion under the evaluated conditions; it is not an independently measured confidence interval or an empirically established 95th-percentile error bound. This sub-millimetre quantity supported successful operation of the evaluated bead-transfer task, which represents the positioning scale relevant to fluid-driven cell-transport applications.

\begin{figure*}[t]
  \centering
  \includegraphics[width=1.0\linewidth]{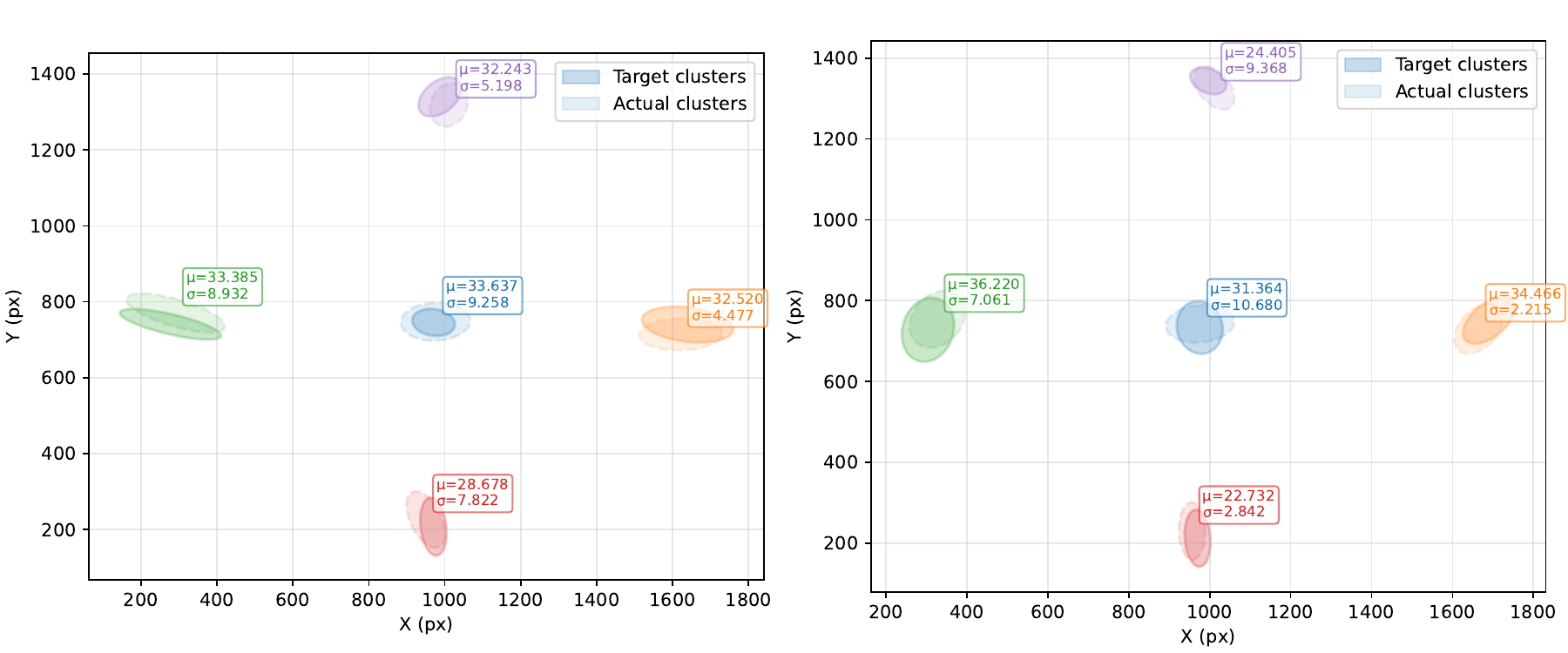}
  \caption{\textbf{Lateral accuracy robustness.} Error ellipses derived from paired Target-Actual clusters. We compare targets reached under Low-Depth (defocused) and High-Depth (focused) conditions. The comparable ellipse sizes ($\sigma \approx 10$\,px) indicate that lateral visual servoing remains precise even under optical blur.}
  \label{fig:lateral-ellipses}
\end{figure*}

\paragraph{Lateral Accuracy under Defocus}
We evaluate lateral precision by analyzing the correspondence between user-specified targets and actual robot positions.
For each target location (e.g., Left Edge), we record the commanded positions across trials as a ``Target Cluster'' and the resulting stabilized robot positions as an ``Actual Cluster''.
These pairs allow us to compute the steady-state lateral error statistics.
Fig.~\ref{fig:lateral-ellipses} visualizes these statistics as error ellipses for two regimes:
\textit{Low Depth} (before depth regulation, defocused) and \textit{High Depth} (after depth regulation, focused).
Despite the reduced image sharpness in the low-depth regime, the lateral dispersion remains comparable to the in-focus case.
The largest observed landing-position dispersion was $\sigma_{\max}^{\mathrm{full}}=10.68$ px in the $1920\times1440$ control-logging coordinates. Using $s_{\mathrm{full}}=2.291\,\mu\mathrm{m}/\mathrm{px}$ gives the dispersion-based engineering estimate:
\[
e_{xy,\mathrm{eng}}
\approx
2s_{\mathrm{full}}\sigma_{\max}^{\mathrm{full}}
=
2\times2.291\times10.68
\approx48.9\,\mu\mathrm{m}.
\]
This dispersion-based estimate ($\sim$50\,$\mu$m) characterizes lateral landing repeatability under the evaluated focused and defocused conditions.

\begin{figure*}[t]
  \centering
  \includegraphics[width=\linewidth]{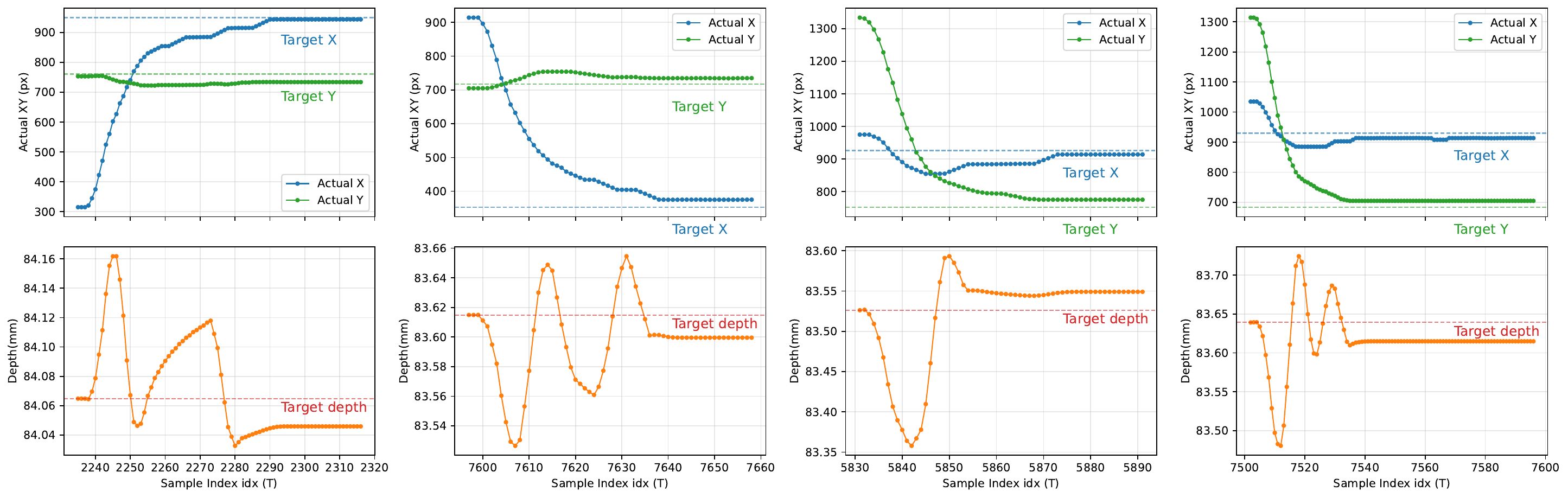}
  \caption{\textbf{Active axial constraint.} Large lateral motions (top row) are achieved with minimal axial disturbance (bottom row), as the controller actively regulates $|e_z| < 30\,\mu\text{m}$, validating the kinematic decoupling.}
  \label{fig:xy-depth-constraint}
\end{figure*}

\paragraph{Decoupling Effectiveness}
Finally, we verify the efficacy of the virtual axial constraint during the High-Depth phase. Due to kinematic coupling, large lateral motions naturally induce axial drift.
Fig.~\ref{fig:xy-depth-constraint} plots the tip trajectory during wide lateral excursions.
The complementary controller successfully suppresses axial drift, maintaining the depth error within a tight band:
\[
|e_z(t)| \le \delta_z \approx 30\,\mu\text{m}.
\]
This active compensation eliminates the need for repeated depth re-focusing after lateral moves, significantly streamlining the manipulation workflow.

\subsection{User Study: Human-Robot Collaboration}
\label{sec:userstudy}

\paragraph{Experimental Design}
We conducted a controlled user study to evaluate task performance and operator workload.
Eight volunteers (3 engineers with experience using the system and 5 novices) were recruited.
To ensure a fair comparison across interaction modes, the operation monitor was spatially aligned with the workspace, ensuring that the visual motion axis aligned with the motor axis. This setup eliminates the cognitive burden of mental rotation (e.g., preventing the ``moving left but seeing right'' conflict).

\paragraph{Task Protocol}
Participants performed a ``pick-and-place'' task using microbeads scattered on the slide. The goal was to approach a target bead, aspirate it, and transfer it to the opposite side of the FoV (e.g., Left $\to$ Right).
The critical challenge lies in the aspiration phase: due to the limited power of the suction pump, the tool tip must be positioned in extremely close proximity to the bead for successful capture, demanding high-precision alignment.

\paragraph{Interaction Conditions}
We compared three methods:
\begin{enumerate}
    \item \textbf{Manual (Freehand):} The operator holds the micropipette directly (detached from the robot) with unrestricted entry points, using a manual button for suction.
    \item \textbf{Steady-hand:} The micropipette is mounted on the robot. The operator guides the robot via the loadcell using admittance control (with identical parameters to the proposed Macro layer), providing active damping but no visual guidance.
    \item \textbf{Agent (Ours):} The proposed shared-control framework. The user provides macro guidance to bring the tool into FoV, clicks on the monitor to designate the lateral target (with the depth target implicitly defined by the current focal plane), and the system autonomously executes micro-alignment via visual servoing.
\end{enumerate}
Each participant completed trials followed by recorded sessions in a counterbalanced order.

\begin{figure}[t]
  \centering
  \includegraphics[width=\linewidth]{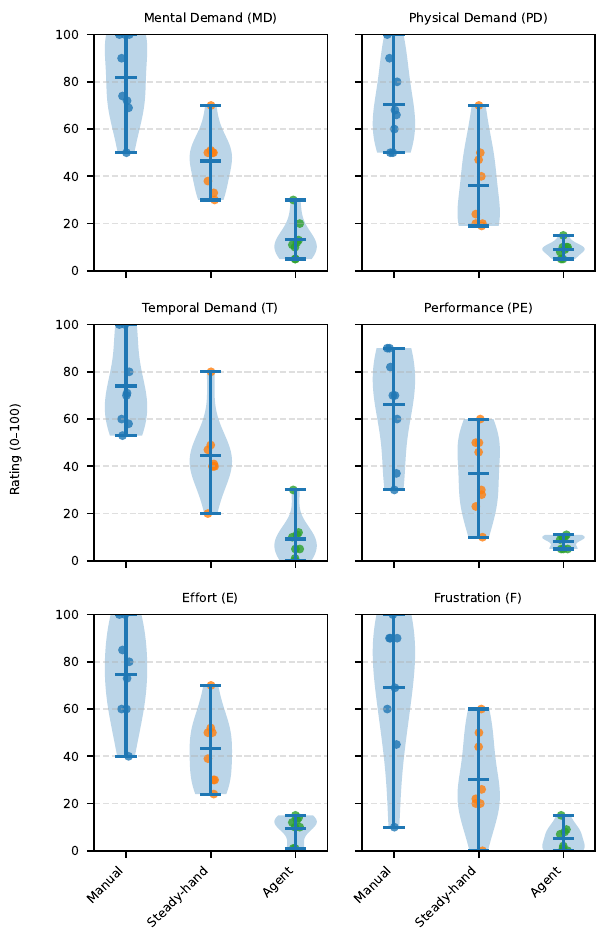}
\caption{NASA-TLX workload distributions ($N=8$). The proposed \textbf{Agent} method achieves the lowest scores across all subscales.  Notably, the lower Performance (PE) score under the Agent condition indicates better perceived performance (Agent $\approx 8$ vs.\ Manual $\approx 66$).}
  \label{figure:nasa_tlx_violins}
\end{figure}

\paragraph{Results and Workload Analysis}
We assessed perceived workload using the NASA-TLX (scale 0--100, where higher indicates higher cost/failure).
Fig.~\ref{figure:nasa_tlx_violins} visualizes the distributions.
The \textbf{overall} (unweighted) TLX score is lowest for \textbf{Agent} ($9.1 \pm 3.0$), intermediate for \textbf{Steady-hand} ($39.7 \pm 12.6$), and highest for \textbf{Manual} ($72.8 \pm 14.9$).

This represents a $\approx 87.5\%$ reduction in workload from Manual to Agent and a $\approx 77.1\%$ reduction in workload from Steady-hand to Agent. 
We note that the NASA-TLX results constitute subjective workload evidence. The completion outcome and absence of observed tip-loss or safety-triggered events are reported separately in Sec.~VI-C as protocol-specific system-level indicators.

Crucially, the \emph{Performance} (PE) subscale, which measures perceived failure or lack of success, provides indicative evidence of our method's effectiveness.
The PE score drops drastically from \textbf{Manual ($66.13 \pm 22.72$)} to \textbf{Steady-hand ($37.13 \pm 16.92$)} and further to \textbf{Agent ($8.13 \pm 2.64$)}.
These scores indicate better perceived performance under the Agent condition during the evaluated task, rather than an objectively measured reduction in task failure.
Similarly, Mental Demand drops from $\sim 81.9$ (Manual) to $\sim 13.3$ (Agent), and Effort from $\sim 74.8$ to $\sim 9.5$, indicating that the automated visual servoing effectively offloads the precision requirements from the human operator.

\begin{table}[t]
\footnotesize
\centering
\caption{Stratified NASA-TLX Overall Scores by Experience Level 
(Mean $\pm$ Std, scale 0--100).}
\label{tab:tlx_stratified}
\begin{tabularx}{\columnwidth}{l Y Y Y}
\toprule
\textbf{Group} & \textbf{Manual} & \textbf{Steady-hand} & \textbf{Agent} \\
\midrule
Engineer ($n=3$) & $60.3 \pm 15.3$ & $34.0 \pm 10.9$ & $7.4 \pm 2.7$ \\
Novice ($n=5$)   & $80.2 \pm 8.1$  & $43.0 \pm 12.4$ & $10.1 \pm 2.7$ \\
\midrule
All ($n=8$)      & $72.8 \pm 14.9$ & $39.7 \pm 12.6$ & $9.1 \pm 3.0$ \\
\bottomrule
\end{tabularx}
\end{table}

To examine whether the Agent condition reduces the subjective workload gap between  experienced and novice operators, we stratified the NASA-TLX results by experience level. Table~V reports the overall TLX scores for each  group and condition. Under Manual operation, novices reported  substantially higher workload than engineers ($80.2 \pm 8.1$ vs.\ $60.3 \pm 15.3$), a gap of $19.9$ points  consistent with the expected skill differential. 
The observed difference between the two experience groups decreased from $19.9$ points under Manual operation to $9.0$ points under Steady-hand control and $2.7$ points under the Agent condition. Both groups showed comparable reductions from Manual to Agent ($87.5\%$ for novices and $87.7\%$ for engineers).  These subgroup results ($n=5$ and $n=3$) suggest that the Agent condition may reduce experience-related differences in subjective workload.

\section{Discussions and Limitations}

Our work re-frames steady-hand co-manipulation not just as a control mode, but as a mechanism to accumulate training data for weakly supervised learning. This approach directly addresses the hierarchy of challenges outlined in the introduction.

\paragraph{Data Scarcity via Weak Supervision}
In Sec.~\ref{sec:intro}, we highlighted the difficulty of establishing consistent 3D supervision for transparent tools. While manual annotation of 2D tip locations is feasible, it fails to capture the axial depth correlations required for 3D control. Our results indicate that short steady-hand segments serve as effective ``weak teachers'' by providing synchronized kinematic-visual streams. This setup leverages the robot's physical motion to automatically generate dense depth labels via focus-defocus correlation---a task practically infeasible for human annotators due to the lack of explicit depth cues.
By distilling these kinematic cues into a heatmap regressor, we achieve landmark precision (PCK@6 $\approx$ 95\%) without frame-by-frame manual supervision.
Crucially, this pipeline is robust to the ``Domain Shift'' challenge: the detector trained on these warm ups generalized well to test sets with distractors (bubbles/particles), balancing FPR/FNR (7.03\%/6.50\%).
These results support the use of warm-up-derived supervision for the evaluated visual conditions, but do not establish generalization to unseen microscope configurations, tool types, or specimens.

The calibration registration error further serves as an automatic pseudo-label quality indicator, yielding an estimated $79.1\%$ reduction in annotation effort for the evaluated dataset and workflow relative to frame-by-frame manual annotation.
(Sec.~V-B, Appendix~III).

\paragraph{Closing the Tip-Regulation Loop under Kinematic Uncertainty}
We identified limited depth observability as a key constraint. Our dual-task observer addresses this by fusing a lateral tracker evaluated on data from two microscope setups with a setup-dependent depth estimator.
The depth estimator is trained on demonstrations of axial scans, while the ``active axial constraint'' strategy mitigates estimation errors by decoupling lateral and axial servoing (as verified in Fig.~\ref{fig:xy-depth-constraint}).

Furthermore, since our error-state estimation operates primarily in the velocity domain, the uncertainty introduced by repeated tool re-installation (e.g., variations in tool length) is effectively mitigated, as absolute positioning is not a prerequisite for the closed loop.
To counterbalance the lack of absolute encoders, we rely on a visibility indicator jointly inferred by the heatmap detector and SoftgateKF; this ensures stable operation as long as the detection-tracking pipeline holds.
Consequently, the task-critical effects of tool-induced geometric uncertainties are compensated within the lateral and axial tip-regulation loops under the evaluated setup.

Regarding the robustness of the perception modules to optical  variability, the lateral detector was trained on data pooled from two distinct microscope setups under varying illumination conditions, and generalised well to challenging test sequences (FPR/FNR: 7.03\%/6.50\%, Table~\ref{tab:unified_results}-A). The depth estimator, whilst requiring a per-setup 
warm-up sequence, does not rely on explicit optical parameters; the defocus-depth mapping is fitted implicitly from kinematic-visual correspondences during warm-up.

\paragraph{Tip Regulation under Geometric Uncertainty}
Simultaneously, uncertainties stemming from non-synchronous latency and hand-eye misalignment are explicitly decomposed and minimized by the calibration module.

Here, the temporal treatment is used as a local refinement within the approximately synchronised motion--vision stream, preventing small frame-level offsets from dominating the spatial calibration objective.

The reported focal-plane alignment supports compensation for task-relevant effects of uncertain tip geometry through relative visual feedback under the evaluated setup: \textit{Robot} $\leftrightarrow$ \textit{Unknown End-Effector} $\leftrightarrow$ \textit{Tip} $\leftrightarrow$ \textit{Focal Plane} $\leftrightarrow$ \textit{Z-axis Module}. 
Under the stated engineering assumptions of the evaluated setup, relative visual feedback closes the task-critical tip-regulation loop without requiring explicit identification of the variable tool length or the absolute focal-plane position. This addresses the dominant tip-installation uncertainty for the reported closed-loop task, rather than constituting a complete metric reconstruction of the optical--robotic kinematic chain.

The residual systematic bias ($\sim 113\,\mu\text{m}$) approaches the physical observability limit, as the $76\,\mu\text{m}$ depth of field (2X objective) creates a significant ``dead zone" where visual cues vanish.

\paragraph{Comparison with Human Physiological Limits}
For engineering context, we construct a coarse reference by combining the $76\,\mu$m depth of field of the present optical setup with the approximately $202\,\mu$m hand-motion amplitude reported for trained eye surgeons during station-keeping~\cite{riviere1997characteristics}:
\[
e_{\mathrm{human,ref}}
\approx \delta_d+\delta_{\mathrm{tremor}}
\approx 76\,\mu\mathrm{m}+202\,\mu\mathrm{m}
=278\,\mu\mathrm{m}.
\]
The system-level axial estimate ($\approx291\,\mu$m) is therefore of the same order of magnitude as this coarse optics-plus-tremor reference. Because the reference combines quantities obtained under different conditions and does not model visual correction or task-specific operator strategies, it is used to indicate the relevant positioning scale rather than to establish direct human--robot performance equivalence.

\paragraph{Closing the Uncertainty Loop}
We observe a tight alignment between our error bounds from dataset and control performance.
For planar motion, the dispersion-based landing-position estimate ($\approx48.9\,\mu$m), computed in the $1920\times1440$ logging coordinates, remained below the tracker's empirical 95th-percentile error reported in the $480\times360$ control coordinates:
\[
8.25\,\mathrm{px}\times
9.164\,\mu\mathrm{m}/\mathrm{px}
\approx75.6\,\mu\mathrm{m}.
\]
The applicable coordinate system and pixel scale are stated explicitly for each conversion.

\paragraph{ Macro-Micro Transition and Confidence-Gated Safeguards}
The experiments demonstrate the operation of the proposed confidence-gated control safeguards under the evaluated protocol.
The seamless transition from admittance-based macro guidance to autonomous micro servoing reduced operator workload by $\sim 77.1\%$.
The ``Visibility Guard'' served as the hierarchical authority manager proposed in the Introduction, allowing autonomous visual-servo commands only when the detection confidence exceeded the predefined threshold. Together with continuous manual override through the admittance-controlled macro mode, this demonstrates the intended safeguard behaviour under the evaluated conditions.

\paragraph{Limitations}
Despite these advances, limitations remain:
\begin{enumerate}
    \item \textbf{Precision Ceiling:} The approximately $49\,\mu$m value characterizes lateral landing dispersion in the evaluated trajectories. Its adequacy for oocyte or embryo handling requires task-specific evaluation~\cite{mendizabal2025digitally}.
    \item \textbf{Actuation Modeling:} We treat fluid dispensing as exogenous. Integrating pressure-flow dynamics is the next step toward fully automating the ``aspirate" phase.
    \item \textbf{Temporal alignment in distributed inference:} The current calibration assumes a centrally controlled acquisition pipeline, or timestamped streams that can be coarsely aligned. The proposed method therefore serves as a local refinement for small residual offsets. For edge-computing or remote inference services with larger timing mismatch, explicit global time-shift estimation would be a natural extension.
    \item \textbf{Generalization constraints:} To prioritize closed-loop bandwidth, we employed aggressive input down-sampling. While effective in our high-contrast microscopy environment, this trade-off may compromise robustness in general surgical scenes with complex textures or low contrast, necessitating higher input resolutions and optimized detection engineering in future iterations. 
\end{enumerate}

\section{Conclusion}

This work presents a weakly supervised framework that uses non-task-specific warm-up trajectories as a unified source of visual--kinematic supervision for tip representation learning, relative depth estimation, markerless registration, and closed-loop control initialisation. This shared supervision mechanism enables 3D tip regulation without requiring task-level demonstrations, external markers, or frame-by-frame manual depth annotation.
Quantitatively,the evaluated trajectories yielded an empirical lateral error of approximately $49\,\mu$m and a conservative axial error of approximately $291\,\mu$m. Although these engineering estimates characterize the reported laboratory observations and support the feasibility of the integrated framework, they do not establish clinically validated positioning accuracy.

From a human--robot interaction perspective, the Agent condition reduced self-reported overall workload by $77.1\%$ relative to steady-hand control, while the descriptive subgroup results suggest reduced experience-related differences in subjective workload.
By exposing an interpretable uncertainty budget, from detection confidence to calibration residuals, our framework offers a promising pathway for laboratory-scale microscope-guided micromanipulation.

\bibliographystyle{IEEEtran}

\bibliography{IEEEabrv,
  BIB_xx-TIE-xxxx,
  ieeebib_settings
  }

\appendices

\newcommand{\nomitem}[2]{\item \(\displaystyle #1\)\;--\;#2}

\ifCLASSOPTIONcaptionsoff
  \newpage
\fi

\vfill
\end{document}


\bstctlcite{IEEEexample:BSTcontrol}
    \title{Learning From a Steady Hand: A Weakly Supervised Agent for Robot Assistance under Microscopy}
  
\ifblind

    \author{Anonymous Authors} 
\else
  \author{Huanyu Tian$^{1}$,
      Martin~Huber$^{1}$,
      Lingyun~Zeng$^{1}$,
      Zhe~Han$^{1}$,
      Wayne~Bennett$^{2}$,
      Giuseppe~Silvestri$^{2}$,  
      Gerardo~Mendizabal-Ruiz$^{2,4}$, 
      Tom~Vercauteren$^{1}$, 
      Alejandro~Chavez-Badiola$^{3,2}$,
      and~Christos~Bergeles$^{1,2}$

\thanks{*This work was supported by Innovate UK under grant agreement 10111748.}
\thanks{$^{1}$H.~Tian, M.~Huber, Z.~Han, L.~Zeng, T.~Vercauteren, and C.~Bergeles are with the School of Biomedical Engineering \& Imaging Sciences, King’s College London, UK. Corresponding author: {\tt\small huanyu.tian@kcl.ac.uk}}    
\thanks{$^{2}$W.~Bennett, G.~Silvestri, G.~Mendizabal-Ruiz, C.~Bergeles, and A.~Chavez-Badiola are with Conceivable Life Sciences, New York City, US and London, UK.}
\thanks{$^{3}$A.~Chavez-Badiola is with Hope IVF Mexico, Mexico.}
\thanks{$^{4}$ G.~Mendizabal-Ruiz is with Departamento de Bioingeniería Traslacional, Universidad de Guadalajara, Guadalajara, Jalisco, Mexico}
}
\fi

\markboth{IEEE TRANSACTIONS ON ROBOTICS (Appendix)
}{Author \MakeLowercase{\textit{et al.}}: }

\maketitle

\newcommand{\nomitem}[2]{\item \(\displaystyle #1\)\;—\;#2}

\section*{Appendix: Nomenclature}

\subsection*{Frames, states, and kinematics}
\begin{itemize}
  \nomitem{\{\mathcal{B}\}, \{\mathcal{E}\}, \{\mathcal{C}\}, \{\mathcal{I}\}}{Robot base, end–effector, camera (optical), and image (pixel) frames.}
  \nomitem{\mathbf{q}\in\mathbb{R}^{n_q},\ \dot{\mathbf{q}}}{Joint vector and joint velocity.}
  \nomitem{\mathbf{p}(\mathbf{q}),\ \mathbf{R}(\mathbf{q})}{End–effector position and rotation in $\{\mathcal{B}\}$.}
  \nomitem{\mathbf{J}(\mathbf{q}),\ \mathbf{J}_p}{Geometric Jacobian and its translational block.}
  \nomitem{\mathbf{w}_t=\big[\mathbf{f}_t^\top\ \boldsymbol{\tau}_t^\top\big]^\top}{Measured wrench (force $\mathbf{f}_t$, torque $\boldsymbol{\tau}_t$).}
  \nomitem{\mathbf{x}_t,\ \mathbf{x}_r}{Target and actual end–effector pose.}
\end{itemize}

\subsection*{Admittance and macro motion}
\begin{itemize}
  \nomitem{\mathbf{v}^{\mathrm{adm}}_t}{Admittance Cartesian linear velocity command.}
  \nomitem{\mathbf{M}_d,\ \mathbf{B}_d}{Virtual inertia and damping (diagonal matrices) in $\mathbf{M}_d\dot{\mathbf{v}}^{\mathrm{adm}}_t + \mathbf{B}_d\mathbf{v}^{\mathrm{adm}}_t=\tilde{\mathbf f}_t$.}
  \nomitem{\mathrm{dz}_\delta(\cdot)}{Dead–zone on force with threshold(s) $\delta$.}
  \nomitem{\mathrm{sat}_{v_{\max}}(\cdot)}{Velocity saturation with limit $v_{\max}$.}
  \nomitem{\Delta t}{Control/sample period.}
\end{itemize}

\subsection*{Imaging model and hand–eye}
\begin{itemize}
  \nomitem{\mathbf{y}=[u,v]^\top,\ \mathbf{c}=[c_x,c_y]^\top}{Pixel coordinates and principal point in $\{\mathcal{I}\}$.}
  \nomitem{\mathbf{p}_{\mathrm{cam}}\in\mathbb{R}^3}{Tip in the camera frame; $\mathbf{t}_{\mathrm{vec}}$ camera translation.}
  \nomitem{\mathbf{L}_{\mathrm{img}}\in\mathbb{R}^{2\times 3}}{Weak–perspective interaction matrix (image Jacobian) with $\mathbf{y}=\mathbf{c}+\mathbf{L}_{\mathrm{img}}\mathbf{p}_{\mathrm{cam}}$.}
  \nomitem{s>0}{Pixel scale (px per world unit) in weak–perspective model.}
  \nomitem{\mathbf{R}\in\mathrm{SO}(3),\ \Theta}{Hand–eye (extrinsic) rotation and its Euler parameterization $\Theta=(\theta_x,\theta_y,\theta_z)$.}
  \nomitem{\mathbf{L}_{\mathrm{rot}}(\mathbf{p}_{\mathrm{cam}})}{Rotational interaction matrix; e.g., $\mathbf{L}_{\mathrm{rot}}=-\,\mathbf{L}_{\mathrm{img}}[\mathbf{p}_{\mathrm{cam}}]_\times$.}
  \nomitem{\mathbf{r}_{\mathrm{tip}}}{Tool–tip offset in $\{\mathcal{E}\}$; $\mathbf{p}^{3D}=\mathbf{R}_{EE}\mathbf{r}_{\mathrm{tip}}+\mathbf{t}_{EE}$.}
\end{itemize}

\subsection*{Perception (planar tip and depth)}
\begin{itemize}
  \nomitem{\hat{\mathbf{p}}^{\mathrm{tip}}_t}{Estimated 2D tip location.}
  \nomitem{H,\ \sigma}{Gaussian tip heatmap and its width (pixels).}
  \nomitem{P^t=C(I_t,\hat{\mathbf{p}}^{\mathrm{tip}}_t,L)}{Tip–centered image patch (crop size $L\times L$).}
  \nomitem{\nabla P^t=(G_x^t,G_y^t)}{Patch gradients (e.g., Sobel) for sharpness.}
  \nomitem{z_t,\ z^\star}{Robot–reported axial position; focal–plane depth estimate.}
  \nomitem{\tilde z=(z-z^\star)/s_z}{Normalized axial offset; $s_z$ scale, $\varepsilon_z$ near–focus margin.}
  \nomitem{\zeta(\tilde z)\in\{\text{below},\text{near},\text{above}\}}{Depth–class label; $\hat{\boldsymbol{\zeta}}$ softmax probability; $f_d$ (regression) and $f_c$ (classification) heads.}
  \nomitem{\tau_{*}}{Soft–gate parameters for lateral/depth filter}
\end{itemize}

\subsection*{Calibration and losses}
\begin{itemize}
  \nomitem{\mathcal{L}_{\mathrm{Chamfer}}}{Bi–directional Chamfer alignment loss (projected 3D tips vs 2D detections).}
  \nomitem{\mathcal{L}_{\mathrm{vel}},\ \lambda_{\mathrm{vel}}\!\ge\!0}{Velocity–consistency loss (fixed $\mathbf{J}_{\mathrm{img}}$) and its weight.}
  \nomitem{\rho_{\mathrm{Huber}}(\cdot;\kappa)}{Huber penalty with scale $\kappa$.}
\end{itemize}

\subsection*{Estimator, filtering, and lifting}
\begin{itemize}
  \nomitem{\mathbf{e}^{(x)}_t}{Planar error state $[e_{x,t},e_{y,t},\dot e_{x,t},\dot e_{y,t}]^\top$.}
  \nomitem{\mathbf{e}^{(z)}_t}{Axial (depth) error state $[e_{z,t},\dot e_{z,t}]^\top$.}
  \nomitem{\mathbf{A}_x,\ \mathbf{A}_z}{Constant–velocity transition matrices.}
  \nomitem{\mathbf{Q}_x,\ \mathbf{Q}_z}{Process–noise covariances.}
  \nomitem{\mathbf{H}^{\mathrm{upd}}_x=\mathbf{I}_2,\ \mathbf{H}^{\mathrm{upd}}_z=1}{Measurement models.}
  \nomitem{\mathcal{R}^{(\chi)}_t}{Score–dependent measurement–noise covariance, $\chi\in\{x,z\}$.}
  \nomitem{\mathcal{K}^{(\chi)}_t}{(Soft–gated) Kalman gain.}
  \nomitem{\mathbf{e}^{(x)}_{\mathrm{task}}}{Planar task–space error $(\mathbf{L}_{\mathrm{img}}\hat{\mathbf{R}})^\dagger (\hat{\mathbf{p}}_{\mathrm{tip}}-\mathbf{p}^\star)$.}
  \nomitem{e^{(z)}_{\mathrm{task}}}{Axial task–space error; $\mathbf{S}_z$ selects control axis.}
\end{itemize}

\subsection*{Conformal gating and reliability}
\begin{itemize}
  \nomitem{d_t^2}{Mahalanobis nonconformity score on innovations.}
  \nomitem{q=\mathcal{Q}_{1-\alpha}(\{d_t^2\})}{$(1-\alpha)$ quantile used as radius/threshold.}
  \nomitem{s^{(x)}_t,\ g^{(x)}_t}{Planar confidence and gate, $g^{(x)}_t=\mathrm{clip}(s^{(x)}_t/\tau_x,0,1)$.}
  \nomitem{\tau_x,\ \sigma^2_{x0},\ \gamma_x}{Map score to $\mathcal{R}^{(x)}_t=\sigma^2_{x0}\exp(-\gamma_x s^{(x)}_t)\mathbf{I}_2$.}
  \nomitem{S^{(z)}_t,\ g^{(z)}_t,\ \tau_z,\ \sigma^2_{z0},\ \gamma_z}{Axial reliability and analogous gating terms.}
\end{itemize}

\subsection*{Micro control and fusion}
\begin{itemize}
  \nomitem{\mathbf{u}^{\mathrm{micro}}_t=[u^{\mathrm{x}}_t,\ u^{\mathrm{z}}_t]^\top}{Micro (lateral/depth) task–space velocity.}
  \nomitem{\mathcal{K}_{\mathrm{x}},\ \mathcal{K}_z}{Feedback gains (e.g., LQR) for lateral/axial regulation.}
  \nomitem{v^{\mathrm{x}}_{\max},\ v^{\mathrm{z}}_{\max}}{Saturation limits for lateral/axial micro motion.}
  \nomitem{\mathbf{J}^\dagger_\lambda}{Damped resolved–rate inverse $\mathbf{W}_q^{-1}\mathbf{J}_p^\top(\mathbf{J}_p \mathbf{W}_q^{-1}\mathbf{J}_p^\top+\lambda^2 \mathbf{I}_3)^{-1}$.}
  \nomitem{\dot{\mathbf{q}}_{\mathrm{macro}},\ \dot{\mathbf{q}}_{\mathrm{micro}}}{Macro and micro joint commands, fused at execution.}
\end{itemize}

\subsection*{Optimization (macro QP)}
\begin{itemize}
  \nomitem{\dot{\mathbf{q}}_{\mathrm{macro}}}{Solution of a single–step QP with tracking and limits.}
  \nomitem{\mathbf{W}_p,\ \lambda_R,\ \lambda_{\dot q}}{Weights for position, rotation regularization, and joint effort.}
\end{itemize}

\subsection*{Data scales and metrics}
\begin{itemize}
  \nomitem{s_{\mathrm{dataset}},\ s_{\mathrm{run}}}{Pixel–to–length scales for dataset/runtime (e.g., $\mu$m/px).}
  \nomitem{\mathrm{PCK}_\delta,\ \tau_{\mathrm{det}},\ \mathrm{Conf}_t}{Keypoint accuracy within $\delta$ px; detection threshold; heatmap confidence.}
\end{itemize}

\subsection*{Datasets (as referenced in text)}
\begin{itemize}
  \nomitem{\mathcal{D}_{\mathrm{tip,train}}}{Tip-training dataset used for hand–eye calibration/learning.}
  \nomitem{\mathcal{D}_{\mathrm{cal}}}{Dataset used for conformal analysis.}
  \nomitem{\mathcal{D}_{\rm depth}}{Dataset used for depth estimator training.}
\end{itemize}


\begin{table*}[t]
\centering
\caption{Controller hyperparameters used in all physical experiments.}
\label{tab:ctrl_hparams}
\begin{tabular}{l l l}
\hline
Symbol & Description & Value (units) \\
\hline
$\mathbf{M}_d$ & Virtual inertia (diagonal) & $\mathrm{diag}(1.0,\,1.0,\,1.0)$ (kg) \\
$\mathbf{B}_d$ & Virtual damping (diagonal) & $\mathrm{diag}(12.0,\,12.0,\,12.0)$ (N\,s/m) \\
$\delta$ & Dead-zone thresholds on force & $[1.0,\ 1.0,\ 1.0]$ (N) \\
$v^{\mathrm{x}}_{\max}$ & Lateral saturation &  $10.0$ (mm/s) \\
$v^{\mathrm{z}}_{\max}$ & Axial saturation & $5.0$ (mm/s) \\
$\Delta t$ & Control period & $33.3$ (ms) \quad ($\approx 30$ Hz) \\
$\mathbf{W}_p$ & Position-tracking weight & $\mathrm{diag}(10^4,\,10^4,\,10^4)$ \\
$\mathbf{W}_q$ & Joint-space weight in damped inverse & $\mathbf{I}_{6}$ \\
$\lambda$ & Damping factor in inverse & $0.1$ \\
$K_{\mathrm{x}}$ & Lateral feedback gain (LQR-like / adaptive) & matrix, state/J-dependent;\ $Q_{\!fb}=1600\,\mathbf{I}_3,\ R_{\!fb}=\mathbf{I}_6$ \\
$K_{z}$ & Axial feedback gain (LQR) & $[\,9.72,\ 2.73\,]$ \\
$\delta_z^{th}$ & Axial tolerance (switching/constraint) & $0.03$ (mm) \\
\hline
\end{tabular}
\end{table*}

\begin{table*}[t]
\centering
\caption{Hyperparameters for calibration losses, fixed Jacobians, and soft gating.}
\label{tab:align_gate}
\begin{tabular}{l l l}
\hline
Symbol & Description & Value \\
\hline
$\lambda_{\mathrm{vel}}$ & Weight of velocity-consistency loss $L_{\mathrm{vel}}$ & 0.10 \\
$\kappa$ & Huber penalty scale in $L_{\mathrm{Chamfer}}$ & 0.12 \\
$q$ & Quantile for conformal radius ($1-\alpha$) & $95\%$ \\
$\tau_x,\ \sigma^2_{x0},\ \gamma_x$ & Planar score$\to$noise mapping & $6.30;\ 4.0;\ 1.0$ \\
$\tau_z,\ \sigma^2_{z0},\ \gamma_z$ & Axial score$\to$noise mapping & $1.0$;\ $0.003^2$;\ $1.0$ \\
\hline
\end{tabular}
\centering
\caption{PGD parameters used in robustness and tracking ablations.}
\label{tab:pgd}
\begin{tabular}{l l l}
\hline
Symbol & Description & Value \\
\hline
$\varepsilon$ & Perturbation budget (e.g., $\ell_\infty$) & $0.02$ \\
$\alpha$ & Step size & $0.03$ \\
$K$ & Number of iterations & $1$ \\
Norm & Attack norm & $\ell_\infty$ \\
Targeting & Attack type & untargeted \\
Eval. cadence & How often perturb frames & every $10$ frames \\
\hline
\end{tabular}
\end{table*}

\section{Stability and Error Analysis of the Soft-Gated Estimator}
\label{app:softgate_stability}

\subsection{Stability Analysis with Watchdog Constraint}
We analyze the stability of the recursive filter defined in Sec.~III. The error covariance update follows the modified Riccati recursion:
\begin{equation}
\mathbf{P}^{(\chi)}_{t} = \big(\mathbf{I}_{n_\chi} - g^{(\chi)}_t \mathbf{K}^{\text{nom}}_t \mathbf{H}_\chi\big)\mathbf{P}^{(\chi)}_{t|t-1},
\label{eq:gated_cov_update}
\end{equation}
where $g^{(\chi)}_t \in [0,1]$ is the scalar reliability gate. 
Intermittent observations can lead to covariance growth during signal loss \cite{Sinopoli2004Kalman}. In the constant-velocity model used here, prolonged prediction without informative updates can likewise increase covariance.

To limit uninterrupted prediction during prolonged occlusions or tracking failures, the system implements a deterministic \textbf{watchdog timer}. Let $t_{last}$ be the timestamp of the last valid update (where $g^{(\chi)}_t > \tau_{\text{safe}}$). The watchdog enforces a reset if $t - t_{last} > T_{\max}$.
Consequently, the maximum open-loop propagation steps $N_{\max} = \lceil T_{\max}/\Delta t \rceil$ is strictly finite. Starting from a finite covariance, prediction over at most $N_{\max}$ steps admits the following finite-horizon bound:
\begin{equation}
\|\mathbf{P}_t\| \le \sum_{k=0}^{N_{\max}} \|\mathbf{F}_\chi\|^k \|\mathbf{Q}_\chi\| \|\mathbf{F}_\chi^\top\|^k + \|\mathbf{F}_\chi\|^{N_{\max}} \|\mathbf{P}_{t_{last}}\| \|\mathbf{F}_\chi^\top\|^{N_{\max}}.
\end{equation}
This bounds covariance growth over the stated prediction horizon. A uniform bound across repeated resets additionally requires a uniformly bounded reset covariance; the finite-horizon expression does not establish BIBO stability or bounded actual estimation error during persistent visual failure.

\subsection{Task-Space Error Mapping}
The controller operates on the estimated task-space error derived from the filter outputs. 
For the planar channel, the sensor observation is the pixel residual $\mathbf{y}^{(x)}_t = \hat{\mathbf{p}}_{\mathrm{tip}, t} - \mathbf{p}^\star$. 
The mapping to task space utilizes the image Jacobian $\mathbf{L}_{\mathrm{img}}$ and the estimated hand-eye rotation $\hat{\mathbf{R}}$:
\begin{equation}
\hat{\mathbf{e}}^{(x)}_{\text{task},t} = (\mathbf{L}_{\mathrm{img}}\hat{\mathbf{R}})^{\dagger}\,(\mathbf{H}_x \hat{\mathbf{e}}^{(x)}_{t}),
\end{equation}
where $\mathbf{H}_x \hat{\mathbf{e}}^{(x)}_{t}$ extracts the position component (first two elements) from the 4D state vector. 
For the axial channel, the scalar depth error is lifted via the control axis selector $\mathbf{S}_z$:
\begin{equation}
\hat{e}^{(z)}_{\text{task},t} = \mathbf{S}_z\,\hat{\mathbf{R}}\,[0,0,1]^\top\,(\mathbf{H}_z \hat{\mathbf{e}}^{(z)}_{t}).
\end{equation}

A first-order perturbation analysis shows that calibration rotation error $\delta\boldsymbol\phi$ and temporal misalignment $\delta t$ introduce bias. 
A first-order engineering estimate of the planar error contributions is:
\[
\|\mathbf{e}^{(x)}_{\text{task}}\| \lesssim s \big(\|e_{\text{det}}\| + \|e_{\text{cal}}\| + \|e_{\text{async}}\|\big),
\]
where $s$ is the pixel-to-metric scale. 
In our configuration, the effective resolution is $480{\times}360$. Taking representative detector and calibration residual magnitudes of approximately $5$ px each gives a combined engineering reference of approximately $90\,\mu\mathrm{m}$, before accounting for any additional temporal-misalignment contribution. The operational gate does not directly guarantee these true error magnitudes.
The observed $3\sigma$ dispersion is $\approx 73.41\,\mu\mathrm{m}$, corresponding to an empirical $2\sigma$-level dispersion of $\approx 49\,\mu\mathrm{m}$. This dispersion is a separate repeatability measure, rather than verification of a theoretical positioning-error bound.

\section{Bounded Commands and Local Stability Considerations}
\label{app:ctrl_sketch}

This section discusses command boundedness and local feedback behavior for the unified macro--micro controller defined in Sec.~\ref{sec:micro_controller}. It does not establish global convergence or input-to-state stability of the complete switched system.

\paragraph{Boundedness of the Command.}
The macro command $\dot{\mathbf{q}}^{\rm macro}_t$ is generated by the QP in \eqref{eq:macro_qp_final}, which explicitly enforces box constraints $\underline{\dot{\mathbf{q}}} \le \dot{\mathbf{q}}^{\rm macro} \le \overline{\dot{\mathbf{q}}}$.
The micro command $\mathbf{u}^{\mathrm{micro}}_t$ is governed by the visibility guard $\mathcal{I}_{\text{vis}}$ and the saturation functions in \eqref{eq:lat_dominant_lat}--\eqref{eq:depth_dominant_dep}, ensuring $\|\mathbf{u}^{\mathrm{micro}}_t\| \le V_{\max}^{\rm micro}$ for some finite bound.
The fused command $\dot{\mathbf{q}}_{\rm cmd} = \dot{\mathbf{q}}^{\rm macro} + \mathbf{J}_p^\dagger{}_\lambda \mathbf{u}^{\mathrm{micro}}$ involves the damped pseudo-inverse $\mathbf{J}_p^\dagger{}_\lambda$, which has norm bounded above by $1/(2\lambda\sqrt{\lambda_{\min}(\mathbf{W}_q)})$.
For fixed $\lambda>0$ and positive-definite $\mathbf{W}_q$, both terms are bounded, so the total command $\dot{\mathbf{q}}_{\rm cmd}$ is bounded. This does not by itself establish closed-loop stability.

\paragraph{Local Feedback Behavior.}
The control law in Sec.~\ref{sec:micro_controller} switches between lateral-dominant and depth-dominant phases.
During lateral-dominant operation, lateral feedback regulates the estimated planar error while the axial virtual fixture compensates for drift from the registered depth.
During depth-dominant operation, lateral micro commands are suppressed and axial feedback regulates the estimated depth error.
The feedback gains are selected using the corresponding local design models. The realized behavior additionally depends on estimation accuracy, calibration residuals, saturation, macro input, and switching.

\paragraph{Scope of the Stability Analysis.}
Stability of individual local design models does not alone establish stability under switching. No common Lyapunov function or switching condition is established here. Accordingly, this analysis establishes bounded commands, while the experiments provide evidence of closed-loop operation under the evaluated protocol.

\section{Experiments}
\label{app:calibration_audit}

\paragraph{Calibration audit}
For each clip \(c\), we use the symmetric Bi-Chamfer registration error as the audit score:
\[
\begin{aligned}
E_{\mathrm{audit}}(c)
=&\;
\frac{1}{2N_c}\sum_{i=1}^{N_c}\min_{j}
\left\|\mathbf{p}_{i}^{\mathrm{proj}}-\mathbf{q}_{j}^{\mathrm{img}}\right\|_2 \\
&+
\frac{1}{2M_c}\sum_{j=1}^{M_c}\min_{i}
\left\|\mathbf{q}_{j}^{\mathrm{img}}-\mathbf{p}_{i}^{\mathrm{proj}}\right\|_2 .
\end{aligned}
\]
where \(\mathbf{p}_{i}^{\mathrm{proj}}\) is the projected robot-tip trajectory after calibration and \(\mathbf{q}_{j}^{\mathrm{img}}\) is the pseudo-labelled image-tip trajectory. Clips are flagged for targeted review when
\[
E_{\mathrm{audit}}(c) > \tau_{\mathrm{audit}},
\qquad
\tau_{\mathrm{audit}} = 2\,\mathrm{median}_{c}\!\left(E_{\mathrm{audit}}(c)\right).
\]

\begin{table}[h]
\footnotesize
\centering
\caption{Per-session calibration audit statistics. 
Threshold: 30.22\,px ($2\times$ median).}
\label{tab:calibration_audit}
\begin{tabularx}{\columnwidth}{l Y Y Y Y Y Y}
\toprule
\textbf{Session} & \textbf{Clips} & \textbf{Auto-approved} & 
\textbf{Flagged} & \textbf{Mean (px)} & \textbf{Min (px)} & 
\textbf{Max (px)} \\
\midrule
S1 &  5 & 5 & 0 &  5.00 &  3.03 & 11.75 \\
S2 &  4 & 3 & 1 & 13.61 &  4.39 & 38.48 \\
S3 &  2 & 2 & 0 &  5.21 &  4.30 &  6.13 \\
S4 & 15 & 14 & 1 & 15.93 &  5.81 & 30.39 \\
S5 &  6 & 5 & 1 & 19.87 &  6.91 & 44.65 \\
S6 & 12 & 9 & 3 & 21.40 &  5.66 & 41.27 \\
S7 &  6 & 6 & 0 & 16.02 &  6.95 & 28.26 \\
S8 &  8 & 7 & 1 & 20.75 &  7.16 & 35.15 \\
\midrule
\textbf{Total} & \textbf{58} & \textbf{51} & \textbf{7} & 
\textbf{16.67} & \textbf{3.03} & \textbf{44.65} \\
\bottomrule
\end{tabularx}
\end{table}

Table~\ref{tab:calibration_audit} reports per-session audit statistics across 58 clips from 8 recording sessions (27,203 frames total, 25,073 valid detections, detection rate 92.2\%). The symmetric Bi-Chamfer error ranged from 3.03 to 44.65\,px (median: 15.11\,px, mean: 16.67\,px, std: 10.42\,px). Clips exceeding a threshold of 30.22\,px ($2\times$ the median) were flagged for targeted human review.

Annotation cost was estimated relative to a frame-by-frame manual annotation baseline at 2\,s/frame. The annotation-effort reduction is computed as
\[
\eta_{\mathrm{save}}
=
1 -
\frac{T_{\mathrm{proposed}}}{T_{\mathrm{manual}}},
\]
where \(T_{\mathrm{manual}}\) is the estimated frame-by-frame annotation time and \(T_{\mathrm{proposed}}\) includes sparse SAM\,2 prompting, automatic propagation, and targeted review of flagged clips. This gives a baseline effort of 15.11\,h for 27,203 frames. The proposed workflow required approximately 3.15\,h, corresponding to an engineering estimate of \(\eta_{\mathrm{save}}=79.1\%\) reduction in annotation effort.

\paragraph{Uncertainty-aware VAE-DEB baseline.}

To examine whether model-level uncertainty can be incorporated directly into the depth-learning objective, we implemented a VAE-style probabilistic variant of the Depth Estimation Block (DEB) as a representative distributional-output baseline. The input remains the same tip-centred crop \(P^t\) used by the deterministic DEB. The backbone first extracts the compact embedding \(\psi_t\). Instead of mapping \(\psi_t\) directly to a point depth estimate, the VAE-DEB predicts a latent Gaussian distribution:
\[
q_{\phi}(\mathbf{h}_t \mid P^t)
=
\mathcal{N}\!\left(
\boldsymbol{\mu}_{\phi}(P^t),
\operatorname{diag}(\boldsymbol{\sigma}^{2}_{\phi}(P^t))
\right),
\]
where \(\mathbf{h}_t\) is the latent variable used for probabilistic depth regression. During training, latent samples are obtained using the reparameterisation trick:
\[
\mathbf{h}_t^{(k)}
=
\boldsymbol{\mu}_{\phi}(P^t)
+
\boldsymbol{\sigma}_{\phi}(P^t) \odot \boldsymbol{\epsilon}^{(k)},
\qquad
\boldsymbol{\epsilon}^{(k)} \sim \mathcal{N}(\mathbf{0}, \mathbf{I}).
\]
Each latent sample is decoded into a relative depth prediction:
\[
\hat{\tilde z}_{t}^{(k)} = f_d(\mathbf{h}_t^{(k)}),
\]
where \(\tilde z_t\) denotes the normalized relative depth label defined in Sec.~III-B. At inference, \(K\) stochastic samples are drawn. The predictive mean is used as the depth estimate:
\[
\bar{\tilde z}_t =
\frac{1}{K}\sum_{k=1}^{K}\hat{\tilde z}_{t}^{(k)} .
\]
The model-level uncertainty is defined as the sampling standard deviation:
\[
u_t =
\sqrt{
\frac{1}{K}
\sum_{k=1}^{K}
\left(
\hat{\tilde z}_{t}^{(k)} - \bar{\tilde z}_t
\right)^2
}.
\]
The absolute depth error is computed as:
\[
e_t = |\bar{\tilde z}_t - \tilde z_t|.
\]
To quantify whether the predicted uncertainty is informative, we compute the Pearson correlation between the uncertainty values \(\{u_t\}_{t=1}^{N}\) and the absolute errors \(\{e_t\}_{t=1}^{N}\):
\[
\rho_{u,e}
=
\frac{
\sum_{t=1}^{N}(u_t-\bar{u})(e_t-\bar{e})
}{
\sqrt{\sum_{t=1}^{N}(u_t-\bar{u})^2}
\sqrt{\sum_{t=1}^{N}(e_t-\bar{e})^2}
},
\]
where \(\bar{u}=\frac{1}{N}\sum_t u_t\) and \(\bar{e}=\frac{1}{N}\sum_t e_t\). A larger positive \(\rho_{u,e}\) indicates that samples assigned higher uncertainty tend to have larger depth error.

The VAE-DEB is trained with the following objective:
\[
\mathcal{L}_{\mathrm{VAE}}
=
\mathcal{L}_{\mathrm{depth}}
+
\lambda_{\mathrm{cls}}\mathcal{L}_{\mathrm{cls}}
+
\beta_{\mathrm{KL}}\mathcal{L}_{\mathrm{KL}}
+
\lambda_{\mathrm{corr}}\mathcal{L}_{\mathrm{corr}},
\]
where \(\mathcal{L}_{\mathrm{depth}}\) is the depth regression loss on \(\bar{\tilde z}_t\), \(\mathcal{L}_{\mathrm{cls}}\) is the auxiliary below/near/above classification loss used in the original DEB, and
\[
\mathcal{L}_{\mathrm{KL}}
=
D_{\mathrm{KL}}
\left(
q_{\phi}(\mathbf{h}_t \mid P^t)
\,\|\, 
\mathcal{N}(\mathbf{0},\mathbf{I})
\right).
\]
For the uncertainty-correlation variant, the additional term is defined over each mini-batch as:
\[
\mathcal{L}_{\mathrm{corr}} = 1 - \rho_{u,e}.
\]
This encourages predicted uncertainty to be positively associated with absolute depth error within each mini-batch. We use this baseline only as an uncertainty-aware comparison; the main pipeline retains the deterministic DEB and propagates reliability at the controller/filtering interface.

\begin{table*}[t]
\centering
\caption{Uncertainty-aware VAE-DEB baseline compared with the deterministic depth baselines from Table~\ref{tab:unified_results}(B). The VAE-DEB row is an additional reviewer-requested probabilistic baseline; its uncertainty--error Pearson/Spearman correlations were \(0.235/0.012\), computed with \(K=30\) Monte-Carlo samples.}
\label{tab:vae_depth_uncertainty}
\begin{tabular}{lccccccc}
\toprule
\multicolumn{8}{c}{\textbf{Depth Estimation and Classification}} \\
\midrule
Model (dim) & MSE & R$^2$ & Cls. Acc. & F1 (macro) & F1 (C0) & F1 (C1) & F1 (C2) \\
\midrule
LDSENet (32D)             & 0.167 & 0.813 & 0.771 & 0.731 & 0.861 & 0.501 & 0.831 \\
ResNet-based (16D)        & \textbf{0.156} & \textbf{0.825} & \textbf{0.793} & \textbf{0.753} & 0.855 & 0.543 & 0.861 \\
FocDepthFormerNet (32D)   & 0.178 & 0.801 & 0.769 & 0.728 & 0.836 & 0.504 & 0.845 \\
\midrule
VAE-DEB + Unc. Corr. (16D)$^\dagger$ & 0.436 & 0.468 & 0.583 & 0.518 & 0.642 & 0.141 & 0.772 \\
\bottomrule
\end{tabular}

\vspace{1mm}
\footnotesize{$^\dagger$ Additional uncertainty-aware baseline added during revision.}
\end{table*}

\paragraph{Results.}
The VAE-DEB baseline was evaluated on the same depth-evaluation subset used for the deterministic depth models in Table~\ref{tab:unified_results}(B). For VAE-DEB, the reported depth estimate is the predictive mean \(\bar{\tilde z}_t\) computed from \(K=30\) stochastic latent samples, and the regression error is measured against the normalized relative depth label \(\tilde z_t\). Compared with the ResNet-based DEB, VAE-DEB showed degraded regression performance (MSE: \(0.156 \rightarrow 0.436\), \(R^2: 0.825 \rightarrow 0.468\)) and classification performance (accuracy: \(0.793 \rightarrow 0.583\), F1 macro: \(0.753 \rightarrow 0.518\)). Its class-wise F1 scores were \(0.642\), \(0.141\), and \(0.772\) for C0, C1, and C2, respectively.

We then evaluated whether the sampling standard deviation \(u_t\) was informative of the absolute depth error \(e_t = |\bar{\tilde z}_t-\tilde z_t|\). The Pearson correlation \(\rho_{u,e}\) between \(\{u_t\}_{t=1}^{N}\) and \(\{e_t\}_{t=1}^{N}\) was \(0.235\), indicating only a weak linear association between the predicted uncertainty and the realized depth error. The corresponding Spearman correlation was \(0.012\), indicating that \(u_t\) provided almost no reliable rank ordering of high-error samples. Thus, although VAE-DEB produced a sampling-based model-level uncertainty score, this score was not sufficiently informative for downstream reliability gating.

These results suggest that this direct VAE-style probabilistic extension is not sufficient, in the present warm-up-supervised setting, to improve either the accuracy of \(\bar{\tilde z}_t\) or the operational usefulness of \(u_t\). We interpret this degradation as likely reflecting the additional stochastic latent modelling and regularisation introduced by VAE-DEB, which increase learning complexity without producing a sufficiently informative uncertainty signal. We therefore retain the lightweight deterministic DEB in the main pipeline and handle reliability through filtering- and controller-level consistency checks.

\vfill
\bibliographystyle{IEEEtran}
\bibliography{IEEEabrv,BIB_xx-TIE-xxxx}